\documentclass{article}
\usepackage{graphicx}

\newcommand{\trsj}[4]{#1}
\newcommand{\trj}[2]{#1}
\newcommand{\erice}[2]{#2}

\long\def\comment#1{} 

\trsj{
\textwidth 6.0in
\textheight 8.5in
\topmargin -0.25in
\oddsidemargin 0.25in
\evensidemargin 0.25in

}{
\textwidth 6.0in
\textheight 8.5in
\topmargin -0.25in
\oddsidemargin 0.25in
\evensidemargin 0.25in

}{
}{
}

\trsj{
 
}{
 
}{
}{
 
}

\newcommand{\sindex}[1]{}

\newcommand{\Fig}[1]{Figure~\ref{#1}} 
\newcommand{\Tab}[1]{Table~\ref{#1}} 
\newcommand{\Eq}[1]{Equation~\ref{#1}} 
\newcommand{\Eqs}[1]{Equations~\ref{#1}} 
\newcommand{\Sec}[1]{Section~\ref{#1}} 

\newcommand{\node}[1]{{\em #1}}

\newcommand{\G}{\Gamma}
\newcommand{\Nij}{N_{ij}}
\newcommand{\Nijk}{N_{ijk}}
\newcommand{\Npij}{\alpha_{ij}} 
\newcommand{\Npijk}{\alpha_{ijk}} 
 
\newcommand{\X}{X}
\newcommand{\x}{x}
\newcommand{\U}{{\bf X}}

\newcommand{\bX}{{\bf X}}
\newcommand{\bx}{{\bf x}}
\newcommand{\bY}{{\bf Y}}
\newcommand{\by}{{\bf y}}
\newcommand{\bZ}{{\bf Z}}
\newcommand{\bz}{{\bf z}}
\newcommand{\bF}{{\bf F}}
\newcommand{\Pa}{{\bf Pa}}
\newcommand{\Pai}{{\bf Pa}_i}
\newcommand{\pai}{{\bf pa}_i}

\newcommand{\Bs}{S}
\newcommand{\Bsc}{S_c}

\newcommand{\Bsone}{S_1}
\newcommand{\Bstwo}{S_2}

\newcommand{\hBs}{S^h}
\newcommand{\hBsc}{S^h_c}

\newcommand{\hBsone}{S^h_1}
\newcommand{\hBstwo}{S^h_2}

\newcommand{\Th}{\mbox{\boldmath $\theta$}}

\newcommand{\Thi}{\Th_i}
\newcommand{\TBs}{\Th_{s}}

\newcommand{\TBsone}{\Th_{s1}}
\newcommand{\TBstwo}{\Th_{s2}}

\newcommand{\ta}[1]{\theta_{#1}}
\newcommand{\tijk}{\theta_{ijk}}
\newcommand{\Tij}{\Th_{ij}}

\newcommand{\sTBs}{\mbox{{\boldmath {\footnotesize $\theta$}}}_s}

\newcommand{\CTBs}{\Theta_{s}}

\trsj{

\begin{document}

\title{A Tutorial on Learning With Bayesian Networks}
\author{
David Heckerman\\ heckerma@hotmail.com}
\date{November 1996 (Revised January 2022)}

\maketitle

}{

\begin{document}

\title{Bayesian Networks for Data Mining}
\author{
David Heckerman\\ Microsoft Research\\ Redmond WA, 98052-6399\\
heckerma@microsoft.com}

\maketitle

}{

\begin{document}
\begin{article}

\begin{kluwerdates}
June 30, 1996
June 30, 1996
June 30, 1996
\end{kluwerdates}

\journame{Knowledge Discovery in Databases}
\volnumber{1}
\issuenumber{1}
\issuemonth{September}
\volyear{1996}
\authorrunninghead{D. Heckerman}
\titlerunninghead{Bayesian Networks for Data Mining}

\title{Bayesian Networks for Data Mining}

\authors{David Heckerman}
\email{heckerma@microsoft.com}
\affil{Microsoft Research, 9S, Redmond, WA 98052-6399}

\editor{Usama Fayyad}
}{ 

\begin{opening}
\title{A TUTORIAL ON LEARNING WITH BAYESIAN NETWORKS}

\author{David Heckerman}
\institute{Microsoft Research, Bldg 9S\\
Redmond WA, 98052-6399\\
heckerma@microsoft.com}

\end{opening}

\begin{document}
}

\trsj{

\begin{abstract}
\noindent 
A Bayesian network is a graphical model that encodes probabilistic
relationships among variables of interest.  When used in conjunction
with statistical techniques, the graphical model has several
advantages for data analysis.  One, because the model encodes
dependencies among all variables, it readily handles situations where
some data entries are missing.  Two, a Bayesian network can be used to
learn causal relationships, and hence can be used to gain
understanding about a problem domain and to predict the consequences
of intervention.  Three, because the model has both a causal and
probabilistic semantics, it is an ideal representation for combining
prior knowledge (which often comes in causal form) and data.  Four,
Bayesian statistical methods in conjunction with Bayesian networks
offer an efficient and principled approach for avoiding the
overfitting of data.  In this paper, we discuss methods for
constructing Bayesian networks from prior knowledge and summarize
Bayesian statistical methods for using data to improve these models.
With regard to the latter task, we describe methods for learning both
the parameters and structure of a Bayesian network, including
techniques for learning with incomplete data.  In addition, we relate
Bayesian-network methods for learning to techniques for supervised and
unsupervised learning.  We illustrate the graphical-modeling
approach using a real-world case study.
\end{abstract}

}{

\begin{abstract}
\noindent 
A Bayesian network is a graphical model that encodes probabilistic
relationships among variables of interest.  When used in conjunction
with statistical techniques, the graphical model has several
advantages for data mining.  One, because the model encodes
dependencies among all variables, it readily handles situations where
some data entries are missing.  Two, a Bayesian network can be used to
learn causal relationships, and hence can be used to gain
understanding about a problem domain and to predict the consequences
of intervention.  Three, because the model has both a causal and
probabilistic semantics, it is an ideal representation for combining
domain knowledge (which often comes in causal form) and data.  Four,
Bayesian statistical methods in conjunction with Bayesian networks
offer an efficient and principled approach for avoiding the
overfitting of data.  In this paper, we discuss methods for
constructing Bayesian networks from domain knowledge and summarize
Bayesian statistical methods for using data to improve these models.
With regard to the latter task, we describe methods for learning both
the parameters and structure of a Bayesian network, including
techniques for learning with incomplete data.  In addition, we relate
Bayesian-network methods for learning to techniques for supervised and
unsupervised learning.  We illustrate the graphical-modeling approach
using a real-world case study.
\end{abstract}

}{

\abstract{ A Bayesian network is a graphical model that encodes
probabilistic relationships among variables of interest.  When used in
conjunction with statistical techniques, the graphical model has
several advantages for data mining.  One, because the model encodes
dependencies among all variables, it readily handles situations where
some data entries are missing.  Two, a Bayesian network can be used to
learn causal relationships, and hence can be used to gain
understanding about a problem domain and to predict the consequences
of intervention.  Three, because the model has both a causal and
probabilistic semantics, it is an ideal representation for combining
domain knowledge (which often comes in causal form) and data.  Four,
Bayesian statistical methods in conjunction with Bayesian networks
offer an efficient and principled approach for avoiding the
overfitting of data.  In this paper, we discuss methods for
constructing Bayesian networks from domain knowledge and summarize
Bayesian statistical methods for using data to improve these models.
With regard to the latter task, we describe methods for learning both
the parameters and structure of a Bayesian network, including
techniques for learning with incomplete data.  In addition, we relate
Bayesian-network methods for learning to techniques for supervised and
unsupervised learning.  We illustrate the graphical-modeling
approach using a real-world case study.

}}{

\begin{abstract}
\noindent 
A Bayesian network is a graphical model that encodes probabilistic
relationships among variables of interest.  When used in conjunction
with statistical techniques, the graphical model has several
advantages for data analysis.  One, because the model encodes
dependencies among all variables, it readily handles situations where
some data entries are missing.  Two, a Bayesian network can be used to
learn causal relationships, and hence can be used to gain
understanding about a problem domain and to predict the consequences
of intervention.  Three, because the model has both a causal and
probabilistic semantics, it is an ideal representation for combining
prior knowledge (which often comes in causal form) and data.  Four,
Bayesian statistical methods in conjunction with Bayesian networks
offer an efficient and principled approach for avoiding the
overfitting of data.  In this paper, we discuss methods for
constructing Bayesian networks from prior knowledge and summarize
Bayesian statistical methods for using data to improve these models.
With regard to the latter task, we describe methods for learning both
the parameters and structure of a Bayesian network, including
techniques for learning with incomplete data.  In addition, we relate
Bayesian-network methods for learning to techniques for supervised and
unsupervised learning.  We illustrate the graphical-modeling
approach using a real-world case study.
\end{abstract}

}

\section{Introduction} \label{sec:intro}

A Bayesian network is a graphical model for probabilistic
relationships among a set of variables.  Over the last decade, the
Bayesian network has become a popular representation for encoding
uncertain expert knowledge in expert systems (Heckerman {\em et al.},
1995a).\nocite{HMW95cacm} More recently, researchers have developed
methods for learning Bayesian networks from data.  The techniques that
have been developed are new and still evolving, but they have been
shown to be remarkably effective for some data-analysis problems.

\trj{In this paper, we provide a tutorial on Bayesian networks and
associated Bayesian techniques for extracting and encoding knowledge
from data.}{In this paper, we provide a tutorial on Bayesian networks
and associated Bayesian techniques for data mining---the process of
extracting knowledge from data.}  There are numerous representations
available for \trj{data analysis}{data mining}, including rule bases,
decision trees, and artificial neural networks; and there are many
techniques for \trj{data analysis}{data mining} such as density
estimation, classification, regression, and clustering.  So what do
Bayesian networks and Bayesian methods have to offer?  There are at
least four answers.

One, Bayesian networks can readily handle incomplete data sets.  For
example, consider a classification or regression problem where two of
the explanatory or input variables are strongly anti-correlated.  This
correlation is not a problem for standard supervised learning
techniques, provided all inputs are measured in every case.  When one
of the inputs is not observed, however, most models will produce an
inaccurate prediction, because they do not encode the correlation
between the input variables.  Bayesian networks offer a natural way to
encode such dependencies.

Two, Bayesian networks allow one to learn about causal relationships.
Learning about causal relationships are important for at least two
reasons.  The process is useful when we are trying to gain
understanding about a problem domain, for example, during exploratory
data analysis.  In addition, knowledge of causal relationships allows
us to make predictions in the presence of interventions.  For example,
a marketing analyst may want to know whether or not it is worthwhile
to increase exposure of a particular advertisement in order to
increase the sales of a product.  To answer this question, the analyst
can determine whether or not the advertisement is a cause for
increased sales, and to what degree.  The use of Bayesian networks
helps to answer such questions even when no experiment about the
effects of increased exposure is available.

Three, Bayesian networks in conjunction with Bayesian statistical
techniques facilitate the combination of domain knowledge and data.
\trj{Anyone who has performed a real-world analysis knows the
importance of prior or domain knowledge, especially when data is
scarce or expensive.  The fact that some commercial systems (i.e.,
expert systems) can be built from prior knowledge alone is a testament
to the power of prior knowledge.}{Anyone who has performed a
real-world data-mining task knows the importance of domain or {\em
prior} knowledge.}  Bayesian networks have a causal semantics that
makes the encoding of causal prior knowledge particularly
straightforward.  In addition, Bayesian networks encode the strength
of causal relationships with probabilities.  Consequently, prior
knowledge and data can be combined with well-studied techniques from
Bayesian statistics.

Four, Bayesian methods in conjunction with Bayesian networks and other
types of models offers an efficient and principled approach for
avoiding the over fitting of data.  \trj{As we shall see, there is no need
to hold out some of the available data for testing.  Using the
Bayesian approach, models can be ``smoothed'' in such a way that all
available data can be used for training.}{}

This tutorial is organized as follows.  In Section~\ref{sec:bayes}, we
discuss the Bayesian interpretation of probability and review methods
from Bayesian statistics for combining prior knowledge with data.  In
\Sec{sec:bn}, we describe Bayesian networks and discuss how they can
be constructed from prior knowledge alone.  In \Sec{sec:inf}, we
discuss algorithms for probabilistic inference in a Bayesian network.
In Sections~\ref{sec:nvar} and \ref{sec:miss}, we show how to learn
the probabilities in a fixed Bayesian-network structure, and describe
techniques for handling incomplete data including Monte-Carlo methods
and the Gaussian approximation.  In Sections~\ref{sec:struct} through
\trj{\ref{sec:eg}}{\ref{sec:priors}}, we show how to learn both the
probabilities and structure of a Bayesian network.  Topics discussed
include methods for assessing priors for Bayesian-network structure
and parameters, and methods for avoiding the overfitting of data
including Monte-Carlo, Laplace, BIC, and MDL approximations.  In
Sections~\ref{sec:sup} and \ref{sec:unsup}, we describe the
relationships between Bayesian-network techniques and methods for
supervised and unsupervised learning.  In \Sec{sec:cause}, we show how
Bayesian networks facilitate the learning of causal relationships.  In
\Sec{sec:app}, we illustrate techniques discussed in the tutorial
using a real-world case study.  In \Sec{sec:point}, we give pointers
to software and additional literature.

\section{The Bayesian Approach to Probability and Statistics} 
\label{sec:bayes}

To understand Bayesian networks and associated
\trj{learning}{data-mining} techniques, it is important to understand
the Bayesian approach to probability and statistics.  In this section,
we provide an introduction to the Bayesian approach for those readers
familiar only with the classical view.

In a nutshell, the Bayesian probability of an event $x$ is a person's
{\em degree of belief} in that event.  Whereas a classical probability
is a physical property of the world (e.g., the probability that a coin
will land heads), a Bayesian probability is a property of the person
who assigns the probability (e.g., your degree of belief that the coin
will land heads).  To keep these two concepts of probability distinct,
we refer to the classical probability of an event as the true or
physical probability of that event, and refer to a degree of belief in
an event as a Bayesian or personal probability.  Alternatively, when
the meaning is clear, we refer to a Bayesian probability simply as a
probability.

One important difference between physical probability and personal
probability is that, to measure the latter, we do not need repeated
trials.  For example, imagine the repeated tosses of a sugar cube onto
a wet surface.  Every time the cube is tossed, its dimensions will
change slightly.  Thus, although the classical statistician has a hard
time measuring the probability that the cube will land with a
particular face up, the Bayesian simply restricts his or her attention
to the next toss, and assigns a probability.  As another example,
consider the question: What is the probability that the Chicago Bulls
will win the championship in 2001?  Here, the classical statistician
must remain silent, whereas the Bayesian can assign a probability (and
perhaps make a bit of money in the process).

One common criticism of the Bayesian definition of probability is that
probabilities seem arbitrary.  Why should degrees of belief satisfy
the rules of probability?  On what scale should probabilities be
measured?  In particular, it makes sense to assign a probability of
one (zero) to an event that will (not) occur, but what probabilities
do we assign to beliefs that are not at the extremes?  Not
surprisingly, these questions have been studied intensely.

With regards to the first question, many researchers have suggested
different sets of properties that should be satisfied by degrees of
belief (e.g., Ramsey 1931, Cox 1946, Good 1950, Savage 1954, DeFinetti
1970)\nocite{Ramsey31,Cox46,Good50,Savage54,DeFinetti70}. It turns out
that each set of properties leads to the same rules: the rules of
probability.  Although each set of properties is in itself compelling,
the fact that different sets all lead to the rules of probability
provides a particularly strong argument for using probability to
measure beliefs.

\begin{figure}
\begin{center}
\leavevmode
\includegraphics[width=1.0in]{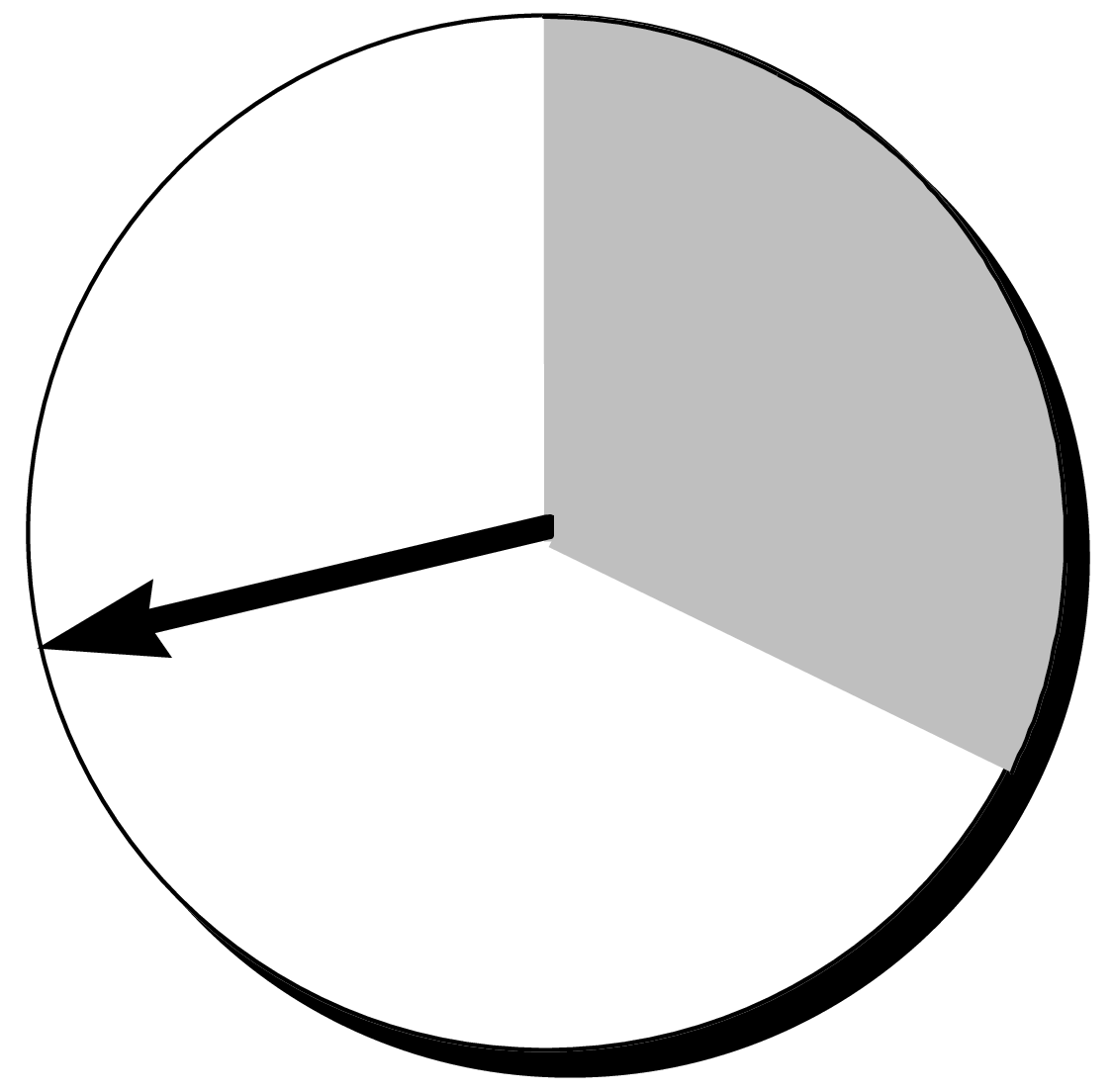}.
\end{center}
\caption{The probability wheel: a tool for assessing probabilities.}
\label{fig:wheel}
\end{figure}

The answer to the question of scale follows from a simple observation:
people find it fairly easy to say that two events are equally likely.
For example, imagine a simplified wheel of fortune having only two
regions (shaded and not shaded), such as the one illustrated in
\Fig{fig:wheel}.  Assuming everything about the wheel as symmetric 
(except for shading), you should conclude that it is equally likely
for the wheel to stop in any one position.  From this judgment and
the sum rule of probability (probabilities of mutually exclusive and
collectively exhaustive sum to one), it follows that your probability
that the wheel will stop in the shaded region is the percent area of
the wheel that is shaded (in this case, 0.3).  

This {\em probability wheel} now provides a reference for measuring
your probabilities of other events.  For example, what is your
probability that Al Gore will run on the Democratic ticket in 2000?
First, ask yourself the question: Is it more likely that Gore will run
or that the wheel when spun will stop in the shaded region?  If you
think that it is more likely that Gore will run, then imagine another
wheel where the shaded region is larger.  If you think that it is more
likely that the wheel will stop in the shaded region, then imagine
another wheel where the shaded region is smaller.  Now, repeat this
process until you think that Gore running and the wheel stopping in
the shaded region are equally likely.  At this point, your probability
that Gore will run is just the percent surface area of the shaded area
on the wheel.

In general, the process of measuring a degree of belief is commonly
referred to as a {\em probability assessment}.  The technique for
assessment that we have just described is one of many available
techniques discussed in the Management Science, Operations Research,
and Psychology literature.  One problem with probability assessment
that is addressed in this literature is that of precision.  Can one
really say that his or her probability for event $x$ is $0.601$ and
not $0.599$?  In most cases, no.  Nonetheless, in most cases,
probabilities are used to make decisions, and these decisions are not
sensitive to small variations in probabilities.  Well-established
practices of {\em sensitivity analysis} help one to know when
additional precision is unnecessary (e.g., Howard and Matheson,
1983)\nocite{BlueBook}.  Another problem with probability assessment
is that of accuracy.  For example, recent experiences or the way a
question is phrased can lead to assessments that do not reflect a
person's true beliefs (Tversky and Kahneman, 1974)\nocite{Tversky74}.
Methods for improving accuracy can be found in the decision-analysis
literature (e.g, Spetzler {\em et al.} (1975)\nocite{Spetzler75}).

Now let us turn to the issue of learning with data.  To illustrate the
Bayesian approach, consider a common thumbtack---one with a round,
flat head that can be found in most supermarkets.  If we throw the
thumbtack up in the air, it will come to rest either on its point
({\em heads}) or on its head ({\em tails}).\footnote{This example is
taken from Howard (1970).\nocite{Howard70a}} Suppose we flip the
thumbtack $N+1$ times, making sure that the physical properties of the
thumbtack and the conditions under which it is flipped remain stable
over time.  From the first $N$ observations, we want to determine the
probability of heads on the $N+1$th toss.

In the classical analysis of this problem, we assert that there is
some physical probability of heads, which is unknown.  We {\em
estimate} this physical probability from the $N$ observations using
criteria such as low bias and low variance.  We then use this estimate as
our probability for heads on the $N+1$th toss.  In the Bayesian
approach, we also assert that there is some physical probability of
heads, but we encode our uncertainty about this physical probability
using (Bayesian) probabilities, and use the rules of probability to
compute our probability of heads on the $N+1$th
toss.\footnote{Strictly speaking, a probability belongs to a single
person, not a collection of people.  Nonetheless, in parts of this
discussion, we refer to ``our'' probability to avoid awkward English.}

To examine the Bayesian analysis of this problem, we need some
notation.  We denote a variable by an upper-case letter (e.g., $X, Y,
X_i, \Theta$), and the state or value of a corresponding variable by
that same letter in lower case (e.g., $x, y, x_i, \theta$).  We denote
a set of variables by a bold-face upper-case letter (e.g., $\bX, \bY,
\bX_i$).  We use a corresponding bold-face lower-case letter (e.g.,
$\bx, \by, \bx_i$) to denote an assignment of state or value to each
variable in a given set.  We say that variable set $\bX$ is in {\em
configuration} $\bx$.  We use $p(X=x|\xi)$ (or $p(x|\xi)$ as a
shorthand) to denote the probability that $X=x$ of a person with state
of information $\xi$.  We also use $p(x|\xi)$ to denote the
probability distribution for $X$ (both mass functions and density
functions).  Whether $p(x|\xi)$ refers to a probability, a probability
density, or a probability distribution will be clear from context.  We
use this notation for probability throughout the paper.  \trj{A summary of
all notation is given at the end of the chapter.}{}

Returning to the thumbtack problem, we define $\Theta$ to be a
variable\footnote{Bayesians typically refer to $\Theta$ as an {\em
uncertain variable}, because the value of $\Theta$ is uncertain.  In
contrast, classical statisticians often refer to $\Theta$ as a {\em
random variable}.  In this text, we refer to $\Theta$ and all
uncertain/random variables simply as variables.}  whose values
$\theta$ correspond to the possible true values of the physical
probability.  We sometimes refer to $\theta$ as a {\em parameter}.  We
express the uncertainty about $\Theta$ using the probability density
function $p(\theta|\xi)$.  In addition, we use $X_l$ to denote the
variable representing the outcome of the $l$th flip, $l=1,\ldots,N+1$,
and $D=\{\X_1=\x_1,\ldots,\X_N=\x_N\}$ to denote the set of our
observations.  Thus, in Bayesian terms, the thumbtack problem reduces
to computing $p(x_{N+1}|D,\xi)$ from $p(\theta|\xi)$.

To do so, we first use Bayes' rule to obtain the probability
distribution for $\Theta$ given $D$ and background knowledge $\xi$:
\begin{equation} \label{eq:bt}
p(\theta|D,\xi) = \frac {p(\theta|\xi) \ p(D|\theta,\xi)}{ p(D|\xi) }
\end{equation}
where 
\begin{equation} \label{eq:ml}
p(D|\xi) = \int p(D|\theta,\xi) \ p(\theta|\xi) \ d\theta
\end{equation}
Next, we expand the term $p(D|\theta,\xi)$.  Both Bayesians and
classical statisticians agree on this term: it is the likelihood
function for binomial sampling.  In particular, given the value of
$\Theta$, the observations in $D$ are mutually independent, and the
probability of heads (tails) on any one observation is $\theta$
($1-\theta$).  Consequently, \Eq{eq:bt} becomes
\begin{equation} \label{eq:bt2}
p(\theta|D,\xi) = \frac{ p(\theta|\xi) \ \theta^h \ (1-\theta)^t }{p(D|\xi)}
\end{equation}
where $h$ and $t$ are the number of heads and tails observed in $D$,
respectively.  The probability distributions $p(\theta|\xi)$ and
$p(\theta|D,\xi)$ are commonly referred to as the {\em prior} and {\em
posterior} for $\Theta$, respectively.  The quantities $h$ and $t$ are
said to be {\em sufficient statistics} for binomial sampling, because they
provide a summarization of the data that is sufficient to compute the
posterior from the prior.  Finally, we average over the possible
values of $\Theta$ (using the expansion rule of probability) to
determine the probability that the $N+1$th toss of the thumbtack will
come up heads:
\begin{eqnarray} \label{eq:bx}
p(\X_{N+1}=heads|D,\xi) & = & \int p(\X_{N+1}=heads|\theta,\xi) \
          p(\theta|D,\xi) \ d\theta \nonumber \\ 
 & = & \int \theta \ p(\theta|D,\xi) \ d\theta 
       \equiv {\rm E}_{p(\theta|D,\xi)}(\theta)
\end{eqnarray}
where ${\rm E}_{p(\theta|D,\xi)}(\theta)$ denotes the expectation of
$\theta$ with respect to the distribution $p(\theta|D,\xi)$.

\begin{figure}
\begin{center}
\leavevmode
\includegraphics[width=4.0in]{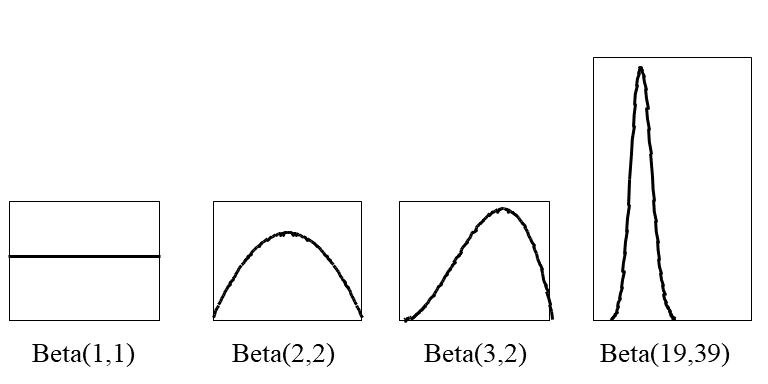}
\end{center}
\caption{Several beta distributions.}
\label{fig:beta}
\end{figure}

To complete the Bayesian story for this example, we need a method to
assess the prior distribution for $\Theta$.  A common approach,
usually adopted for convenience, is to assume that this distribution
is a {\em beta} distribution:
\begin{equation} \label{eq:beta}
p(\theta|\xi) = {\rm Beta}(\theta|\alpha_h,\alpha_t) \equiv
  \frac{\G(\alpha)}{\G(\alpha_h) \G(\alpha_t)} \theta^{\alpha_h - 1}
  (1-\theta)^{\alpha_t-1}
\end{equation}
where $\alpha_h>0$ and $\alpha_t>0$ are the parameters of the beta
distribution, $\alpha=\alpha_h+\alpha_t$, and $\Gamma(\cdot)$ is the
{\em Gamma} function which satisfies $\Gamma(x+1)= x \Gamma(x)$ and
$\Gamma(1)=1$.  The quantities $\alpha_h$ and $\alpha_t$ are often
referred to as {\em hyperparameters} to distinguish them from the
parameter $\theta$.  The hyperparameters $\alpha_h$ and $\alpha_t$
must be greater than zero so that the distribution can be normalized.
Examples of beta distributions are shown in \Fig{fig:beta}.

The beta prior is convenient for several reasons.  By
\Eq{eq:bt2}, the posterior distribution will also be a beta
distribution:
\begin{equation} \label{eq:beta-p}
p(\theta|D,\xi) = 
  \frac{\G(\alpha+N)}{\G(\alpha_h+h) \G(\alpha_t+t)}
  \theta^{\alpha_h + h - 1} (1-\theta)^{\alpha_t + t - 1} = {\rm
  Beta}(\theta|\alpha_h+h,\alpha_t+t)
\end{equation}
We say that the set of beta distributions is a {\em conjugate family
of distributions} for binomial sampling.  Also, the expectation of
$\theta$ with respect to this distribution has a simple form:
\begin{equation} \label{eq:beta-mean}
\int \theta \ {\rm Beta}(\theta|\alpha_h,\alpha_t) \ d\theta
   = \frac{\alpha_h}{\alpha}
\end{equation}
Hence, given a beta prior, we have a simple expression for the
probability of heads in the $N+1$th toss:
\begin{equation} \label{eq:pxb}
p(\X_{N+1}=heads|D,\xi) = \frac{\alpha_h+h}{\alpha+N}
\end{equation}

Assuming $p(\theta|\xi)$ is a beta distribution, it can be assessed in
a number of ways.  For example, we can assess our probability for
heads in the first toss of the thumbtack (e.g., using a probability
wheel).  Next, we can imagine having seen the outcomes of $k$ flips,
and reassess our probability for heads in the next toss.  From
\Eq{eq:pxb}, we have (for $k=1$)
\[
p(X_1=heads|\xi) = \frac{\alpha_h}{\alpha_h+\alpha_t}
\ \ \ \ \ 
p(X_2=heads|X_1=heads,\xi) =
  \frac{\alpha_h+1}{\alpha_h+\alpha_t+1}
\]
Given these probabilities, we can solve for $\alpha_h$ and
$\alpha_t$.  This assessment technique is known as the method of {\em
imagined future data}.  

Another assessment method is based on \Eq{eq:beta-p}.  This equation
says that, if we start with a Beta$(0,0)$ prior\footnote{Technically,
the hyperparameters of this prior should be small positive numbers
so that $p(\theta|\xi)$ can be normalized.}
and observe $\alpha_h$ heads and $\alpha_t$ tails, then our posterior
(i.e., new prior) will be a Beta$(\alpha_h,\alpha_t)$ distribution.
Recognizing that a Beta$(0,0)$ prior encodes a state of minimum
information, we can assess $\alpha_h$ and $\alpha_t$ by determining
the (possibly fractional) number of observations of heads and tails
that is equivalent to our actual knowledge about flipping thumbtacks.
Alternatively, we can assess $p(X_1=heads|\xi)$ and $\alpha$, which
can be regarded as an {\em equivalent sample size} for our current
knowledge.  This technique is known as the method of {\em equivalent
samples}.  Other techniques for assessing beta distributions are
discussed by Winkler (1967)\nocite{Winkler67a} and Chaloner and Duncan
(1983)\nocite{CD83}.

Although the beta prior is convenient, it is not accurate for some
problems.  For example, suppose we think that the thumbtack may have
been purchased at a magic shop.  In this case, a more appropriate
prior may be a mixture of beta distributions---for example,
\[
p(\theta|\xi) = 
  0.4 \ {\rm Beta}(20,1) +
  0.4 \ {\rm Beta}(1,20) +
  0.2 \ {\rm Beta}(2,2)
\]
where 0.4 is our probability that the thumbtack is heavily weighted
toward heads (tails).  In effect, we have introduced an additional
{\em hidden} or unobserved variable $H$, whose states correspond
to the three possibilities: (1) thumbtack is biased toward heads, (2)
thumbtack is biased toward tails, and (3) thumbtack is normal; and we
have asserted that $\theta$ conditioned on each state of $H$ is a beta
distribution.  In general, there are simple methods (e.g., the method
of imagined future data) for determining whether or not a beta prior
is an accurate reflection of one's beliefs.  In those cases where the
beta prior is inaccurate, an accurate prior can often be assessed by
introducing additional hidden variables, as in this example.

So far, we have only considered observations drawn from a binomial
distribution.  In general, observations may be drawn from any physical
probability distribution:
\[
p(x|\Th,\xi) = f(x,\Th)
\]
where $f(x,\Th)$ is the likelihood function with parameters $\Th$.
For purposes of this discussion, we assume that the number of
parameters is finite.  As an example, $X$ may be a continuous
variable and have a Gaussian physical probability distribution with
mean $\mu$ and variance $v$:
\[
p(\x|\Th,\xi) = (2 \pi v)^{-1/2} \ e^{-(\x-\mu)^2/2v}
\]
where $\Th=\{\mu,v\}$.

Regardless of the functional form, we can learn about the parameters
given data using the Bayesian approach.  As we have done in the
binomial case, we define variables corresponding to the unknown
parameters, assign priors to these variables, and use Bayes' rule to
update our beliefs about these parameters given data:
\begin{equation} \label{eq:post-t}
p(\Th|D,\xi) = \frac{p(D|\Th,\xi) \ p(\Th|\xi)}{p(D|\xi)}
\end{equation}
We then average over the possible values of $\Theta$ to make
predictions.  For example,
\begin{equation} \label{eq:post-x}
p(x_{N+1}|D,\xi) = \int p(x_{N+1}|\Th,\xi) \ p(\Th|D,\xi) \ d\Th
\end{equation}
For a class of distributions known as the {\em exponential family},
these computations can be done efficiently and in closed
form.\footnote{Recent advances in Monte-Carlo methods have made it
possible to work efficiently with many distributions outside the
exponential family.  See, for example, Gilks et
al. (1996)\nocite{GRS96}.}  Members of this class include the
binomial, multinomial, normal, Gamma, Poisson, and multivariate-normal
distributions.  Each member of this family has sufficient statistics
that are of fixed dimension for any random sample, and a simple
conjugate prior.\footnote{In fact, except for a few,
well-characterized exceptions, the exponential family is the only
class of distributions that have sufficient statistics of fixed
dimension (Koopman, 1936; Pitman, 1936\nocite{Koopman36,Pitman36}).}
Bernardo and Smith (pp. 436--442, 1994)\nocite{BS94} have compiled the
important quantities and Bayesian computations for commonly used
members of the exponential family.  Here, we summarize these items for
multinomial sampling, which we use to illustrate many of the ideas in
this paper.

In multinomial sampling, the observed variable $X$ is discrete, having
$r$ possible states $x^1,\ldots,x^r$.  The likelihood function is
given by
\[
p(X=x^k|\Th,\xi) = \ta{k}, \ \ \ \ k=1,\ldots,r
\]
where $\Th=\{\ta{2},\ldots,\ta{r}\}$ are the parameters.  (The
parameter $\ta{1}$ is given by $1-\sum_{k=2}^r \ta{k}$.)  In this case,
as in the case of binomial sampling, the parameters correspond to
physical probabilities.  The sufficient statistics for data set
$D=\{X_1=x_1,\ldots,X_N=x_N\}$ are $\{N_1,\ldots,N_r\}$, where $N_i$ is
the number of times $X=x^k$ in $D$.  The simple conjugate prior used
with multinomial sampling is the Dirichlet distribution:
\begin{equation} \label{eq:dir}
p(\Th|\xi) = {\rm Dir}(\Th|\alpha_1,\ldots,\alpha_r) \equiv
  \frac{\G(\alpha)}{\prod_{k=1}^r \G(\alpha_k)} 
    \prod_{k=1}^r \ta{k}^{\alpha_k - 1}
\end{equation}
where $\alpha = \sum_{i=1}^r \alpha_k$, and $\alpha_k >0,
k=1,\ldots,r$.  The posterior distribution $p(\Th|D,\xi)= {\rm
Dir}(\Th|\alpha_1+N_1,\ldots,\alpha_r+N_r)$.  Techniques for assessing
the beta distribution, including the methods of imagined future data
and equivalent samples, can also be used to assess Dirichlet
distributions.  Given this conjugate prior and data set $D$, the
probability distribution for the next observation is given by
\begin{equation} \label{eq:dir-mean}
p(X_{N+1}=x^k|D,\xi) = \int \ta{k} \ {\rm Dir}(\Th|\alpha_1+N_1,\ldots,
  \alpha_r+N_r) \ d\Th = \frac{\alpha_k+N_k}{\alpha+N}
\end{equation}
As we shall see, another important quantity in Bayesian analysis is the
{\em marginal likelihood} or {\em evidence} $p(D|\xi)$.  In this case,
we have
\begin{equation} \label{eq:dir-ml}
p(D|\xi) = \frac{\Gamma(\alpha)}{\Gamma(\alpha+N)} \cdot
  \prod_{k=1}^{r} \frac{\Gamma(\alpha_{k} + N_{k})}{\Gamma(\alpha_{k})}  
\end{equation}  
We note that the explicit mention of the state of knowledge $\xi$ is
useful, because it reinforces the notion that probabilities are
subjective.  Nonetheless, once this concept is firmly in place, the
notation simply adds clutter.  In the remainder of this tutorial, we
shall not mention $\xi$ explicitly.

In closing this section, we emphasize that, although the Bayesian and
classical approaches may sometimes yield the same prediction, they are
fundamentally different methods for learning from data.  As an
illustration, let us revisit the thumbtack problem.  Here, the
Bayesian ``estimate'' for the physical probability of heads is
obtained in a manner that is essentially the opposite of the classical
approach.  

Namely, in the classical approach, $\theta$ is fixed (albeit unknown),
and we imagine all data sets of size $N$ that {\em may be} generated
by sampling from the binomial distribution determined by $\theta$.
Each data set $D$ will occur with some probability $p(D|\theta)$ and
will produce an estimate $\theta^*(D)$.  To evaluate an estimator, we
compute the expectation and variance of the estimate with respect to all
such data sets:
\begin{eqnarray} \label{eq:exp-c}
{\rm E}_{p(D|\theta)}(\theta^*) & = & 
     \sum_D p(D|\theta) \ \theta^*(D) \nonumber \\
{\rm Var}_{p(D|\theta)}(\theta^*) & = & \sum_D p(D|\theta) \ (\theta^*(D) -
  {\rm E}_{p(D|\theta)}(\theta^*))^2
\end{eqnarray}
We then choose an estimator that somehow balances the bias
($\theta-{\rm E}_{p(D|\theta)}(\theta^*)$) and variance of these estimates
over the possible values for $\theta$.\footnote{Low bias and variance
are not the only desirable properties of an estimator.  Other
desirable properties include consistency and robustness.}  Finally, we
apply this estimator to the data set that we actually observe.  A
commonly-used estimator is the maximum-likelihood (ML) estimator,
which selects the value of $\theta$ that maximizes the likelihood
$p(D|\theta)$.  For binomial sampling, we have
\[
\theta^*_{\rm ML}(D) = \frac{N_k}{\sum_{k=1}^r N_k}
\]
\trj{For this (and other types) of sampling, the ML estimator is {\em
unbiased}.  That is, for all values of $\theta$, the ML estimator has
zero bias.  In addition, for all values of $\theta$, the variance of
the ML estimator is no greater than that of any other unbiased
estimator (see, e.g., Schervish, 1995)\nocite{Schervish95}.}{}

In contrast, in the Bayesian approach, $D$ is fixed, and we imagine
all possible values of $\theta$ from which this data set {\em could
have been} generated.  Given $\theta$, the ``estimate'' of the
physical probability of heads is just $\theta$ itself.  Nonetheless,
we are uncertain about $\theta$, and so our final estimate is the
expectation of $\theta$ with respect to our posterior beliefs about
its value:
\begin{equation} \label{eq:exp-b}
{\rm E}_{p(\theta|D,\xi)}(\theta) = \int \theta \ p(\theta|D,\xi) \ d\theta
\end{equation}

The expectations in Equations~\ref{eq:exp-c} and \ref{eq:exp-b} are
different and, in many cases, lead to different ``estimates''.  One way
to frame this difference is to say that the classical and Bayesian
approaches have different definitions for what it means to be a good
estimator.  Both solutions are ``correct'' in that they are self
consistent.  Unfortunately, both methods have their drawbacks, which
has lead to endless debates about the merit of each approach.  For
example, Bayesians argue that it does not make sense to consider the
expectations in \Eq{eq:exp-c}, because we only see a single data set.
If we saw more than one data set, we should combine them into one
larger data set.  In contrast, classical statisticians argue that
sufficiently accurate priors can not be assessed in many situations.
The common view that seems to be emerging is that one should use
whatever method that is most sensible for the task at hand.  We share
this view, although we also believe that the Bayesian approach has
been under used, especially in light of its advantages mentioned in
the introduction (points three and four).  Consequently, in this
paper, we concentrate on the Bayesian approach.

\section{Bayesian Networks} \label{sec:bn}

So far, we have considered only simple problems with one or a few
variables.  In real \trj{learning}{data-mining} problems, however, we
are typically interested in looking for relationships among a large
number of variables.  The Bayesian network is a representation suited
to this task.  It is a graphical model that efficiently encodes the
joint probability distribution (physical or Bayesian) for a large set
of variables.  In this section, we define a Bayesian network and show
how one can be constructed from prior knowledge.

A Bayesian network for a set of variables $\bX=\{X_1,\ldots,X_n\}$
consists of (1) a network structure $S$ that encodes a set of
conditional independence assertions about variables in $\bX$, and (2)
a set $P$ of local probability distributions associated with each
variable.  Together, these components define the joint probability
distribution for $\bX$.  The network structure $S$ is a directed
acyclic graph.  The nodes in $S$ are in one-to-one correspondence with
the variables $\bX$.  We use $X_i$ to denote both the variable and its
corresponding node, and $\Pai$ to denote the parents of node $X_i$ in
$S$ as well as the variables corresponding to those parents.  The {\em
lack} of possible arcs in $S$ encode conditional independencies.  In
particular, given structure $S$, the joint probability distribution
for $\bX$ is given by
\begin{equation} \label{eq:bn-def}
p(\bx) = \prod_{i=1}^n p(x_i|\pai)
\end{equation}
The local probability distributions $P$ are the distributions
corresponding to the terms in the product of \Eq{eq:bn-def}.
Consequently, the pair $(S,P)$ encodes the joint distribution
$p(\bx)$.

The probabilities encoded by a Bayesian network may be Bayesian or
physical.  When building Bayesian networks from prior knowledge alone,
the probabilities will be Bayesian.  When learning these networks from
data, the probabilities will be physical (and their values may be
uncertain).  In subsequent sections, we describe how we can learn the
structure and probabilities of a Bayesian network from data.  In the
remainder of this section, we explore the construction of Bayesian
networks from prior knowledge.  As we shall see in \Sec{sec:priors},
this procedure can be useful in learning Bayesian networks as well.

\begin{figure}
\begin{center}
\leavevmode
\includegraphics[width=4.0in]{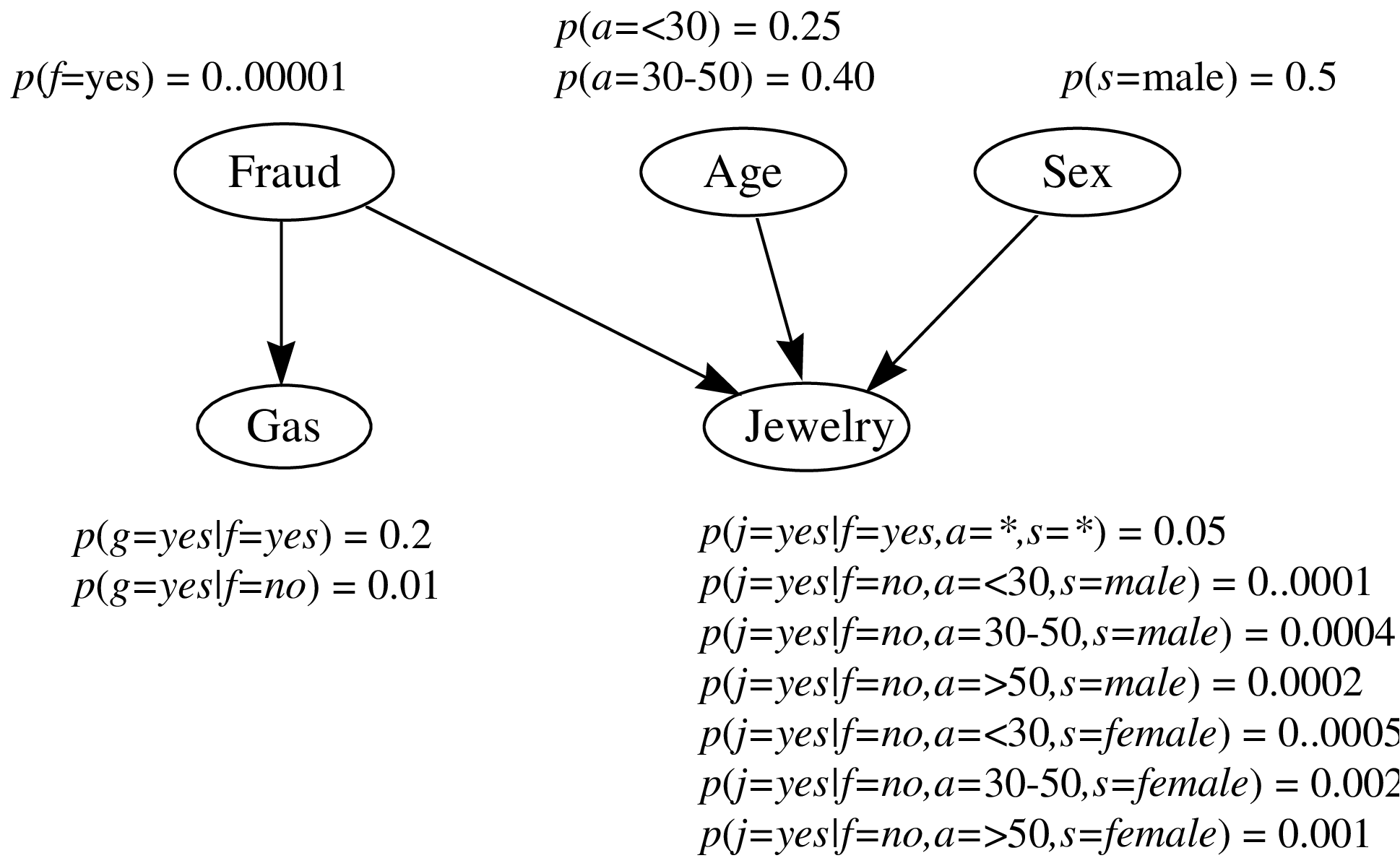}
\end{center}
\caption{A Bayesian-network for detecting credit-card fraud.
Arcs are drawn from cause to effect.  The local probability
distribution(s) associated with a node are shown adjacent to the
node.  An asterisk is a shorthand for ``any state.''  }
\label{fig:bn}
\end{figure}

To illustrate the process of building a Bayesian network, consider the
problem of detecting credit-card fraud.  We begin by determining the
variables to model.  One possible choice of variables for our problem
is \node{Fraud} ($F$), \node{Gas} ($G$), \node{Jewelry} ($J$),
\node{Age} ($A$), and \node{Sex} ($S$), representing whether or not
the current purchase is fraudulent, whether or not there was a gas
purchase in the last 24 hours, whether or not there was a jewelry
purchase in the last 24 hours, and the age and sex of the card holder,
respectively.  The states of these variables are shown in
\Fig{fig:bn}.  Of course, in a realistic problem, we would include
many more variables.  Also, we could model the states of one or more
of these variables at a finer level of detail.  For example, we could
let \node{Age} be a continuous variable.

This initial task is not always straightforward.  As part of this task
we must (1) correctly identify the goals of modeling (e.g., prediction
versus explanation versus exploration), (2) identify many possible
observations that may be relevant to the problem, (3) determine what
subset of those observations is worthwhile to model, and (4) organize
the observations into variables having mutually exclusive and
collectively exhaustive states.  Difficulties here are not unique to
modeling with Bayesian networks, but rather are common to most
approaches.  Although there are no clean solutions, some guidance
is offered by decision analysts (e.g., Howard and Matheson,
1983\nocite{BlueBook}) and (when data are available) statisticians
(e.g., Tukey, 1977\nocite{Tukey77}).

In the next phase of Bayesian-network construction, we build a
directed acyclic graph that encodes assertions of conditional
independence.  One approach for doing so is based on the following
observations.  From the chain rule of probability, we have
\begin{equation} \label{eq:chain}
p(\bx) = \prod_{i=1}^n p(\x_i|\x_1,\ldots,\x_{i-1})
\end{equation}
Now, for every $X_i$, there will be some subset $\Pi_i \subseteq
\{\X_1,\ldots,\X_{i-1}\}$ such that $\X_i$ and
$\{\X_1,\ldots,\X_{i-1}\} \setminus \Pi_i$ are conditionally
independent given $\Pi_i$.  That is, for any $\bx$,
\begin{equation} \label{eq:i-pi}
p(\x_i|\x_1,\ldots,\x_{i-1}) = p(\x_i|\pi_i)
\end{equation}
Combining \Eqs{eq:chain} and \ref{eq:i-pi}, we obtain
\begin{equation} \label{eq:bn-def2}
p(\bx) = \prod_{i=1}^n p(x_i|\pi_i)
\end{equation}
Comparing \Eqs{eq:bn-def} and \ref{eq:bn-def2}, we see that the
variables sets $(\Pi_1,\ldots,\Pi_n)$ correspond to the
Bayesian-network parents $(\Pa_1,\ldots,\Pa_n)$, which in turn fully
specify the arcs in the network structure $S$.

Consequently, to determine the structure of a Bayesian network we (1)
order the variables somehow, and (2) determine the variables sets that
satisfy \Eq{eq:i-pi} for $i=1,\ldots,n$.  In our example, using the
ordering $(F,A,S,G,J)$, we have the conditional
independencies
\begin{eqnarray} \label{eq:car-ci}
p(a|f) & = & p(a)    \nonumber \\
p(s|f,a) & = & p(s)  \nonumber \\
p(g|f,a,s) & = & p(g|f) \nonumber \\
p(j|f,a,s,g) & = & p(j|f,a,s)
\end{eqnarray}
Thus, we obtain the structure shown in \Fig{fig:bn}.

This approach has a serious drawback.  If we choose the variable order
carelessly, the resulting network structure may fail to reveal many
conditional independencies among the variables.  For example, if we
construct a Bayesian network for the fraud problem using the ordering
$(J,G,S,A,F)$, we obtain a fully connected network structure.  Thus,
in the worst case, we have to explore $n!$ variable orderings to find
the best one.  Fortunately, there is another technique for
constructing Bayesian networks that does not require an ordering.  The
approach is based on two observations: (1) people can often readily
assert causal relationships among variables, and (2) causal
relationships typically correspond to assertions of conditional
dependence.  In particular, to construct a Bayesian network for a
given set of variables, we simply draw arcs from cause variables to
their immediate effects.  In almost all cases, doing so results in a
network structure that satisfies the definition $\Eq{eq:bn-def}$.  For
example, given the assertions that \node{Fraud} is a direct cause of
\node{Gas}, and \node{Fraud}, \node{Age}, and \node{Sex} are direct
causes of \node{Jewelry}, we obtain the network structure in Figure
\ref{fig:bn}.  The causal semantics of Bayesian networks are in large
part responsible for the success of Bayesian networks as a
representation for expert systems (Heckerman {\em et al.},
1995a).\nocite{HMW95cacm} In \Sec{sec:cause}, we will see how to learn
causal relationships from data using these causal semantics.

In the final step of constructing a Bayesian network, we assess the
local probability distribution(s) $p(\x_i|\pai)$.  In our fraud
example, where all variables are discrete, we assess one distribution
for $X_i$ for every configuration of $\Pai$.  Example distributions
are shown in \Fig{fig:bn}.

Note that, although we have described these construction steps as a
simple sequence, they are often intermingled in practice.  For
example, judgments of conditional independence and/or cause and effect
can influence problem formulation.  Also, assessments of probability
can lead to changes in the network structure.  Exercises that help one
gain familiarity with the practice of building Bayesian networks can
be found in Jensen (1996)\nocite{Jensen96}.

\section{Inference in a Bayesian Network} \label{sec:inf}

Once we have constructed a Bayesian network (from prior knowledge,
data, or a combination), we usually need to determine various
probabilities of interest from the model.  For example, in our problem
concerning fraud detection, we want to know the probability of fraud
given observations of the other variables.  This probability is not
stored directly in the model, and hence needs to be computed.  In
general, the computation of a probability of interest given a model is
known as {\em probabilistic inference}.  In this section we describe
probabilistic inference in Bayesian networks.

Because a Bayesian network for $\bX$ determines a joint probability
distribution for $\bX$, we can---in principle---use the Bayesian
network to compute any probability of interest.  For example, from the
Bayesian network in \Fig{fig:bn}, the probability of fraud given
observations of the other variables can be computed as follows:
\begin{equation} \label{eq:dumb}
p(f|a,s,g,j) = 
  \frac{p(f,a,s,g,j)}{p(a,s,g,j)} =
  \frac{p(f,a,s,g,j)}{\sum_{f'} p(f',a,s,g,j)}
\end{equation}
For problems with many variables, however, this direct approach is not
practical.  Fortunately, at least when all variables are discrete, we
can exploit the conditional independencies encoded in a Bayesian
network to make this computation more efficient.  In our example, given
the conditional independencies in \Eq{eq:car-ci}, \Eq{eq:dumb} becomes
\begin{eqnarray} \label{eq:smart} 
p(f|a,s,g,j) & = & \frac{p(f)p(a)p(s)p(g|f)p(j|f,a,s)}{\sum_{f'}
  p(f')p(a)p(s)p(g|f')p(j|f',a,s)} \\
& = &
\frac{p(f)p(g|f)p(j|f,a,s)}{\sum_{f'} p(f')p(g|f')p(j|f',a,s)} \nonumber
\end{eqnarray}

Several researchers have developed probabilistic inference algorithms
for Bayesian networks with discrete variables that exploit conditional
independence roughly as we have described, although with different
twists.  For example, Howard and Matheson (1981)\nocite{HM81},
Olmsted (1983)\nocite{Olmsted83}, and Shachter
(1988)\nocite{Shachter87a} developed an algorithm that reverses arcs
in the network structure until the answer to the given probabilistic
query can be read directly from the graph.  In this algorithm, each
arc reversal corresponds to an application of Bayes' theorem. Pearl
(1986) \nocite{Pearl86b} developed a message-passing scheme that
updates the probability distributions for each node in a Bayesian
network in response to observations of one or more variables.
Lauritzen and Spiegelhalter (1988)\nocite{Lauritzen88}, Jensen et
al. (1990)\nocite{JLO90}, and Dawid (1992)\nocite{Dawid92} created an
algorithm that first transforms the Bayesian network into a tree where
each node in the tree corresponds to a subset of variables in $\bX$.
The algorithm then exploits several mathematical properties of this
tree to perform probabilistic inference.  Most recently, D'Ambrosio
(1991) \nocite{DAMbrosio91} developed an inference algorithm that
simplifies sums and products symbolically, as in the transformation
from \Eq{eq:dumb} to \ref{eq:smart}.  The most commonly used algorithm
for discrete variables is that of Lauritzen and Spiegelhalter (1988),
Jensen et al (1990), and Dawid (1992).

\trsj{}{}{}{\begin{sloppypar}}
Methods for exact inference in Bayesian networks that encode
multivariate-Gaussian or Gaussian-mixture distributions have been
developed by Shachter and Kenley (1989)\nocite{Shachter89b} and
Lauritzen (1992),\nocite{Lauritzen92} respectively.  These
methods also use assertions of conditional independence to simplify
inference.  Approximate methods for inference in Bayesian networks
with other distributions, such as the generalized linear-regression
model, have also been developed (Saul {\em et al.}, 1996\nocite{SJJ96};
Jaakkola and Jordan, 1996\nocite{JJ96}).
\trsj{}{}{}{\end{sloppypar}}

Although we use conditional independence to simplify probabilistic
inference, exact inference in an arbitrary Bayesian network for
discrete variables is NP-hard (Cooper, 1990)\nocite{Cooper90b}.  Even
approximate inference (for example, Monte-Carlo methods) is NP-hard
(Dagum and Luby, 1993)\nocite{DL93ai}.  The source of the difficulty
lies in undirected cycles in the Bayesian-network structure---cycles
in the structure where we ignore the directionality of the arcs.  (If
we add an arc from \node{Age} to \node{Gas} in the network structure
of \Fig{fig:bn}, then we obtain a structure with one undirected cycle:
$F-G-A-J-F$.)  When a Bayesian-network structure contains many
undirected cycles, inference is intractable.  For many applications,
however, structures are simple enough (or can be simplified
sufficiently without sacrificing much accuracy) so that inference is
efficient.  For those applications where generic inference methods are
impractical, researchers are developing techniques that are custom
tailored to particular network topologies (Heckerman 1989; Suermondt
and Cooper, 1991; Saul {\em et al.}, 1996; Jaakkola and Jordan,
1996)\nocite{Heckerman89,Suermondt90,SJJ96,JJ96} or to particular
inference queries (Ramamurthi and Agogino, 1988; Shachter et~al.,
1990; Jensen and Andersen, 1990; Darwiche and Provan,
1996)\nocite{Agogino88,Shachter90,Jensen90,DP96}.

\section{Learning Probabilities in a Bayesian Network} \label{sec:nvar}

In the next several sections, we show how to refine the structure and
local probability distributions of a Bayesian network given data.  The
result is set of techniques for \trj{data analysis}{data mining} that
combines prior knowledge with data to produce improved knowledge.  In
this section, we consider the simplest version of this problem: using
data to update the probabilities of a given Bayesian network
structure.

Recall that, in the thumbtack problem, we do not learn the probability
of heads.  Instead, we update our posterior distribution for the
variable that represents the physical probability of heads.  We
follow the same approach for probabilities in a Bayesian network.  In
particular, we assume---perhaps from causal knowledge about the
problem---that the physical joint probability distribution for $\U$
can be encoded in some network structure $\Bs$.  We write
\begin{equation} \label{eq:bn-like}
p(\bx|\TBs,\hBs) = \prod_{i=1}^n p(\x_i|\pai,\Thi,\hBs)
\end{equation} 
where $\Thi$ is the vector of parameters for the distribution
$p(\x_i|\pai,\Thi,\hBs)$, $\TBs$ is the vector of parameters
$(\Th_1,\ldots,\Th_n)$, and $\hBs$ denotes the event (or
``hypothesis'' in statistics nomenclature) that the physical joint
probability distribution can be factored according to
$\Bs$.\footnote{As defined here, network-structure hypotheses overlap.
For example, given $\bX=\{X_1,X_2\}$, any joint distribution for $\bX$
that can be factored according the network structure containing no
arc, can also be factored according to the network structure $X_1
\longrightarrow X_2$.  Such overlap presents problems for model
averaging, described in \Sec{sec:struct}.  Therefore, we should add
conditions to the definition to insure no overlap.  Heckerman and
Geiger (1996)\nocite{HG96tr} describe one such set of conditions.}  In
addition, we assume that we have a random sample
$D=\{\bx_1,\ldots,\bx_N\}$ from the physical joint probability
distribution of $\bX$.  We refer to an element $\bx_l$ of $D$ as a
{\em case}.  As in \Sec{sec:bayes}, we encode our uncertainty about
the parameters $\TBs$ by defining a (vector-valued) variable $\CTBs$,
and assessing a prior probability density function $p(\TBs|\hBs)$.
The problem of learning probabilities in a Bayesian network can now be
stated simply: Given a random sample $D$, compute the posterior
distribution $p(\TBs|D,\hBs)$.

We refer to the distribution $p(\x_i|\pai,\Thi,\hBs)$, viewed as a
function of $\Thi$, as a {\em local distribution function}.  Readers
familiar with methods for supervised learning will recognize that a
local distribution function is nothing more than a probabilistic
classification or regression function.  Thus, a Bayesian network can
be viewed as a collection of probabilistic classification/regression
models, organized by conditional-independence relationships.  Examples
of classification/regression models that produce probabilistic outputs
include linear regression, generalized linear regression,
probabilistic neural networks (e.g., MacKay, 1992a,
1992b)\nocite{MacKay92a,MacKay92b}, probabilistic decision trees
(e.g., Buntine, 1993; Friedman and Goldszmidt,
1996\nocite{FG96}),\nocite{Buntine93} kernel density estimation
methods (Book, 1994)\nocite{Silverman86}, and dictionary methods
(Friedman, 1995).\nocite{Friedman95} In principle, any of these forms
can be used to learn probabilities in a Bayesian network; and, in most
cases, Bayesian techniques for learning are available.  Nonetheless,
the most studied models include the unrestricted multinomial
distribution (e.g., Cooper and Herskovits, 1992), linear regression
with Gaussian noise (e.g., Buntine, 1994; Heckerman and Geiger, 1996),
and generalized linear regression (e.g., MacKay, 1992a and 1992b;
Neal, 1993; and Saul {\em et al.}, 1996).\nocite{MacKay92a,MacKay92b,Neal93}

In this tutorial, we illustrate the basic ideas for learning
probabilities (and structure) using the unrestricted multinomial
distribution.  In this case, each variable $\X_i \in \U$ is discrete,
having $r_i$ possible values $\x^1_i,\ldots,\x^{r_i}_i$, and each
local distribution function is collection of multinomial
distributions, one distribution for each configuration of $\Pai$.
Namely, we assume
\begin{equation} \label{eq:disc-i}
p(\x^k_i|\pai^j,\Thi,\hBs) = \tijk > 0
\end{equation}
where $\pai^1,\ldots,\pai^{q_i}$ ($q_i = \prod_{\X_i \in \Pai} r_i$)
denote the configurations of $\Pai$, and
$\Th_i=((\tijk)_{k=2}^{r_i})_{j=1}^{q_i}$ are the parameters.  (The
parameter $\theta_{ij1}$ is given by $1-\sum_{k=2}^{r_i} \tijk$.)  For
convenience, we define the vector of parameters
\[
\Tij = (\theta_{ij2},\ldots,\theta_{ijr_i})
\]
for all $i$ and $j$.  We use the term ``unrestricted'' to contrast
this distribution with multinomial distributions that are
low-dimensional functions of $\Pai$---for example, the generalized
linear-regression model.

Given this class of local distribution functions, we can compute the
posterior distribution $p(\TBs|D,\hBs)$ efficiently and in closed form
under two assumptions.  The first assumption is that there are no
missing data in the random sample $D$.  We say that the random sample
$D$ is {\em complete}.  The second assumption is that the parameter
vectors $\Tij$ are mutually independent.\footnote{The computation is
also straightforward if two or more parameters are equal.  For
details, see Thiesson (1995).\nocite{Thiesson95}} That is,
\[
p(\TBs|\hBs) = \prod_{i=1}^n \prod_{j=1}^{q_i} p(\Tij|\hBs)
\]
We refer to this assumption, which was introduced by Spiegelhalter and
Lauritzen (1990)\nocite{SL90}, as {\em parameter independence}.  

\trj{
Given that the joint physical probability distribution factors
according to some network structure $\Bs$, the assumption of parameter
independence can itself be represented by a larger Bayesian-network
structure.  For example, the network structure in \Fig{fig:pi}
represents the assumption of parameter independence for $\bX=\{X,Y\}$
($X$, $Y$ binary) and the hypothesis that the network structure $X
\rightarrow Y$ encodes the physical joint probability distribution for
$\bX$.

\begin{figure}
\begin{center}
\leavevmode
\includegraphics[width=2.5in]{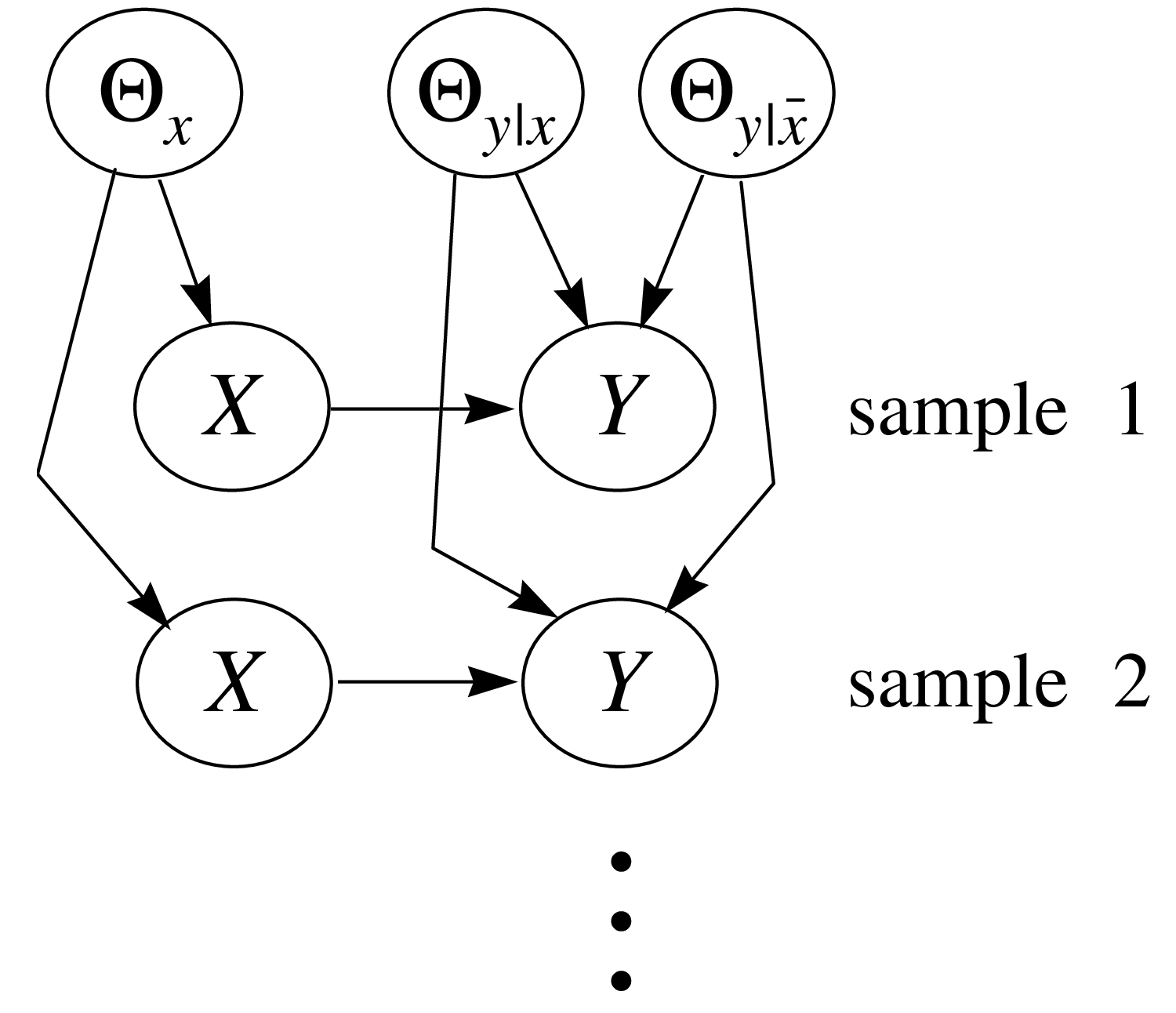}
\end{center}
\caption{A Bayesian-network structure depicting 
the assumption of parameter independence for learning the parameters
of the network structure $X \rightarrow Y$.  Both variables $X$ and
$Y$ are binary.  We use $x$ and $\bar{x}$ to denote the two states of
$X$, and $y$ and $\bar{y}$ to denote the two states of $Y$.}
\label{fig:pi}
\end{figure}
}{}

Under the assumptions of complete data and parameter independence, the
parameters remain independent given a random sample:
\begin{equation} \label{eq:tbs-ind}
p(\TBs|D,\hBs) = \prod_{i=1}^n \prod_{j=1}^{q_i} p(\Tij|D,\hBs)
\end{equation}
Thus, we can update each vector of parameters $\Tij$ independently,
just as in the one-variable case.  Assuming each vector $\Tij$ has the
prior distribution Dir$(\Tij|\alpha_{ij1},\ldots,\alpha_{ijr_i})$, 
we obtain the posterior distribution
\begin{equation} \label{eq:dir-post-n}
p(\Tij|D,\hBs) = {\rm Dir}(\Tij|\alpha_{ij1}+N_{ij1},\ldots,
  \alpha_{ijr_i}+N_{ijr_i})
\end{equation} 
where $N_{ijk}$ is the number of cases in $D$ in which $\X_i=\x_i^k$
and $\Pai=\pai^j$.

As in the thumbtack example, we can average over the possible
configurations of $\TBs$ to obtain predictions of interest.  For
example, let us compute $p(\bx_{N+1}|D,\hBs)$, where $\bx_{N+1}$ is
the next case to be seen after $D$.  Suppose that, in case
$\bx_{N+1}$, $\X_i=\x_i^k$ and $\Pai=\pai^j$, where $k$ and $j$ depend
on $i$.  Thus,
\[
p(\bx_{N+1}|D,\hBs) = {\rm E}_{p(\sTBs|D,\hBs)} 
  \left( \prod_{i=1}^n \tijk \right)
\]
To compute this expectation, we first use the fact that the parameters
remain independent given $D$:
\[
p(\bx_{N+1}|D,\hBs) 
  = \int \prod_{i=1}^n \tijk \ p(\TBs|D,\hBs) \ d\TBs
  = \prod_{i=1}^n \int \tijk \ p(\Tij|D,\hBs) \ d\Tij
\]
Then, we use \Eq{eq:dir-mean} to obtain
\begin{equation} \label{eq:nextc}
p(\bx_{N+1}|D,\hBs) = \prod_{i=1}^n \frac{\Npijk+\Nijk}{\Npij+\Nij}
\end{equation}
where $\Npij = \sum_{k=1}^{r_i} \Npijk$ and $\Nij = \sum_{k=1}^{r_i}
\Nijk$.

These computations are simple because the unrestricted multinomial
distributions are in the exponential family.  Computations for linear
regression with Gaussian noise are equally straightforward (Buntine,
1994; Heckerman and Geiger, 1996).\nocite{HG96tr}

\section{Methods for Incomplete Data} \label{sec:miss}

Let us now discuss methods for learning about parameters when the
random sample is incomplete (i.e., some variables in some cases are
not observed).  An important distinction concerning missing data is
whether or not the absence of an observation is dependent on the
actual states of the variables.  For example, a missing datum in a
drug study may indicate that a patient became too sick---perhaps due
to the side effects of the drug---to continue in the study.  In
contrast, if a variable is hidden (i.e., never observed in any
case), then the absence of this data is independent of state.
Although Bayesian methods and graphical models are suited to the
analysis of both situations, methods for handling missing data where
absence is independent of state are simpler than those where absence
and state are dependent.  In this tutorial, we concentrate on the
simpler situation only.  Readers interested in the more complicated
case should see Rubin (1978), Robins (1986), and Pearl
(1995)\nocite{Rubin78,Robins86,Pearl95bm}.

Continuing with our example using unrestricted multinomial
distributions, suppose we observe a single incomplete case.  Let
$\bY \subset \bX$ and $\bZ \subset \bX$ denote the observed and
unobserved variables in the case, respectively.  Under the
assumption of parameter independence, we can compute the posterior
distribution of $\Tij$ for network structure $\Bs$ as follows:
\begin{eqnarray} \label{eq:miss1}
\erice{
p(\Tij|\by,\hBs) 
& = & \sum_{\bz} p(\bz|\by,\hBs) \ p(\Tij|\by,\bz,\hBs)  \\
& = & (1-p(\pai^j|\by,\hBs)) \left\{ p(\Tij|\hBs) \right\} + \nonumber \\
&&      \sum_{k=1}^{r_i} 
       p(\x_i^k,\pai^j|\by,\hBs) 
       \left\{ p(\Tij|\x_i^k,\pai^j,\hBs) \right\} \nonumber
}{
\lefteqn{p(\Tij|\by,\hBs) 
= \sum_{\bz} p(\bz|\by,\hBs) \ p(\Tij|\by,\bz,\hBs)}  \\
&& = (1-p(\pai^j|\by,\hBs)) \left\{ p(\Tij|\hBs) \right\} + 
      \sum_{k=1}^{r_i} 
       p(\x_i^k,\pai^j|\by,\hBs) 
       \left\{ p(\Tij|\x_i^k,\pai^j,\hBs) \right\} \nonumber
}
\end{eqnarray}
(See Spiegelhalter and Lauritzen (1990)\nocite{SL90} for a
derivation.)  Each term in curly brackets in \Eq{eq:miss1} is a
Dirichlet distribution.  Thus, unless both $\X_i$ and all the
variables in $\Pai$ are observed in case $\by$, the posterior
distribution of $\Tij$ will be a linear combination of Dirichlet
distributions---that is, a Dirichlet mixture with mixing coefficients
$(1-p(\pai^j|\by,\hBs))$ and $p(\x_i^k,\pai^j|\by,\hBs),
k=1,\ldots,r_i$.

When we observe a second incomplete case, some or all of the Dirichlet
components in \Eq{eq:miss1} will again split into Dirichlet mixtures.
That is, the posterior distribution for $\Tij$ we become a mixture of
Dirichlet mixtures.  As we continue to observe incomplete cases, each
missing values for $\bZ$, the posterior distribution for $\Tij$ will
contain a number of components that is exponential in the number of
cases.  In general, for any interesting set of local likelihoods and
priors, the exact computation of the posterior distribution for
$\TBs$ will be intractable.  Thus, we require an approximation for
incomplete data.

\subsection{Monte-Carlo Methods} \label{sec:gibbs}

One class of approximations is based on Monte-Carlo or sampling
methods.  These approximations can be extremely accurate, provided one
is willing to wait long enough for the computations to converge.  

In this section, we discuss one of many Monte-Carlo methods known as
{\em Gibbs sampling}, introduced by Geman and Geman
(1984)\nocite{GG84}.  Given variables $\bX=\{\X_1,\ldots,\X_n\}$ with
some joint distribution $p(\bx)$, we can use a Gibbs sampler to
approximate the expectation of a function $f(\bx)$ with respect to
$p(\bx)$ as follows.  First, we choose an initial state for each of
the variables in $\bX$ somehow (e.g., at random).  Next, we pick some
variable $\X_i$, unassign its current state, and compute its
probability distribution given the states of the other $n-1$
variables.  Then, we sample a state for $\X_i$ based on this
probability distribution, and compute $f(\bx)$.  Finally, we iterate
the previous two steps, keeping track of the average value of
$f(\bx)$.  In the limit, as the number of cases approach infinity,
this average is equal to ${\rm E}_{p(\bx)}(f(\bx))$ provided two conditions
are met.  First, the Gibbs sampler must be {\em irreducible}: The
probability distribution $p(\bx)$ must be such that we can eventually
sample any possible configuration of $\bX$ given any possible initial
configuration of $\bX$.  For example, if $p(\bx)$ contains no zero
probabilities, then the Gibbs sampler will be irreducible.  Second,
each $\X_i$ must be chosen infinitely often.  In practice, an
algorithm for deterministically rotating through the variables is
typically used.  Introductions to Gibbs sampling and other Monte-Carlo
methods---including methods for initialization and a discussion of
convergence---are given by Neal (1993)\nocite{Neal93} and Madigan and
York (1995)\nocite{MY95}.

To illustrate Gibbs sampling, let us approximate the probability
density $p(\TBs|D,\hBs)$ for some particular configuration of $\TBs$,
given an incomplete data set $D=\{\by_1,\ldots,\by_N\}$ and a Bayesian
network for discrete variables with independent Dirichlet priors.  To
approximate $p(\TBs|D,\hBs)$, we first initialize the states of the
unobserved variables in each case somehow.  As a result, we have a
complete random sample $D_c$.  Second, we choose some variable
$\X_{il}$ (variable $\X_i$ in case $l$) that is not observed in the
original random sample $D$, and reassign its state according to the
probability distribution
\[
p(\x'_{il}|D_c \setminus \x_{il},\hBs) = 
  \frac{p(\x'_{il},D_c \setminus \x_{il}|\hBs)}{
        \sum_{\x''_{il}} p(\x''_{il}, D_c \setminus \x_{il}|\hBs)}
\]
where $D_c \setminus \x_{il}$ denotes the data set $D_c$ with
observation $\x_{il}$ removed, and the sum in the denominator runs
over all states of variable $\X_{il}$.  As we shall see in
\Sec{sec:struct}, the terms in the numerator and denominator can be
computed efficiently (see \Eq{eq:bd}).  Third, we repeat this
reassignment for all unobserved variables in $D$, producing a new
complete random sample $D'_c$.  Fourth, we compute the posterior
density $p(\TBs|D'_c,\hBs)$ as described in \Eqs{eq:tbs-ind} and
\ref{eq:dir-post-n}.  Finally, we iterate the previous three steps,
and use the average of $p(\TBs|D'_c,\hBs)$ as our approximation.

\subsection{The Gaussian Approximation} \label{sec:laplace-p}

Monte-Carlo methods yield accurate results, but they are often
intractable---for example, when the sample size is large.  Another
approximation that is more efficient than Monte-Carlo methods and
often accurate for relatively large samples is the {\em Gaussian
approximation} (e.g., Kass {\em et al.}, 1988; Kass and Raftery,
1995)\nocite{KTK88,KR95}.

The idea behind this approximation is that, for large amounts of data,
$p(\TBs|D,\hBs)$ \linebreak $\propto p(D|\TBs,\hBs) \cdot
p(\TBs|\hBs)$ can often be approximated as a multivariate-Gaussian
distribution.  In particular, let
\begin{equation} \label{eq:g-def}
g(\TBs) \equiv \log ( p(D|\TBs,\hBs) \cdot p(\TBs|\hBs) )
\end{equation}
Also, define $\tilde{\TBs}$ to be the configuration of $\TBs$ that
maximizes $g(\TBs)$.  This configuration also maximizes
$p(\TBs|D,\hBs)$, and is known as the {\em maximum a posteriori} (MAP)
configuration of $\TBs$.  Using a second degree Taylor polynomial of
$g(\TBs)$ about the $\tilde{\TBs}$ to approximate $g(\TBs)$, we obtain
\begin{equation} \label{eq:taylor}
g(\TBs) \approx g(\tilde{\TBs}) -\frac{1}{2}
  (\TBs-\tilde{\TBs}) A (\TBs-\tilde{\TBs})^t
\end{equation}
where $(\TBs-\tilde{\TBs})^t$ is the transpose of row vector
$(\TBs-\tilde{\TBs})$, and $A$ is the negative Hessian of $g(\TBs)$
evaluated at $\tilde{\TBs}$.  Raising $g(\TBs)$ to the power of $e$
and using \Eq{eq:g-def}, we obtain
\begin{eqnarray} \label{eq:post-gauss}
p(\TBs|D,\hBs) & \propto &
 p(D|\TBs,\hBs) \ p(\TBs|\hBs) \\
 & \approx &
 p(D|\tilde{\TBs},\hBs) \ p(\tilde{\TBs}|\hBs) 
 \ \exp\{ -\frac{1}{2} (\TBs-\tilde{\TBs}) A (\TBs-\tilde{\TBs})^t \} 
\nonumber
\end{eqnarray}
Hence, $p(\TBs|D,\hBs)$ is approximately Gaussian.

To compute the Gaussian approximation, we must compute $\tilde{\TBs}$
as well as the negative Hessian of $g(\TBs)$ evaluated at
$\tilde{\TBs}$.  In the following section, we discuss methods for
finding $\tilde{\TBs}$.  Meng and Rubin (1991)\nocite{MR91} describe a
numerical technique for computing the second derivatives.  Raftery
(1995)\nocite{Raftery95} shows how to approximate the Hessian using
likelihood-ratio tests that are available in many statistical
packages.  Thiesson (1995)\nocite{Thiesson95} demonstrates that, for
unrestricted multinomial distributions, the second derivatives can be
computed using Bayesian-network inference.

\subsection{The MAP and ML Approximations and the EM Algorithm}
\label{sec:em}

As the sample size of the data increases, the Gaussian peak will
become sharper, tending to a delta function at the MAP configuration
$\tilde{\TBs}$.  In this limit, we do not need to compute averages or
expectations.  Instead, we simply make predictions based on the MAP
configuration.

A further approximation is based on the observation that, as the
sample size increases, the effect of the prior $p(\TBs|\hBs)$
diminishes.  Thus, we can approximate $\tilde{\TBs}$ by the maximum
{\em maximum likelihood} (ML) configuration of $\TBs$:
\[
\hat{\TBs} = \arg\max_{\TBs} \left\{ p(D|\TBs,\hBs) \right\}
\]
 
One class of techniques for finding a ML or MAP is gradient-based
optimization.  For example, we can use gradient ascent, where we
follow the derivatives of $g(\TBs)$ or the likelihood $p(D|\TBs,\hBs)$
to a local maximum.  Russell {\em et al.} (1995)\nocite{RBKK95} and Thiesson
(1995)\nocite{Thiesson95} show how to compute the derivatives of the
likelihood for a Bayesian network with unrestricted multinomial
distributions.  Buntine (1994) discusses the more general case where
the likelihood function comes from the exponential family.  Of course,
these gradient-based methods find only local maxima.

Another technique for finding a local ML or MAP is the
expectation--maximization (EM) algorithm (Dempster {\em et al.},
1977).\nocite{Dempster77} To find a local MAP or ML, we begin by
assigning a configuration to $\TBs$ somehow (e.g., at random).  Next,
we compute the {\em expected sufficient statistics} for a complete
data set, where expectation is taken with respect to the joint
distribution for $\bX$ conditioned on the assigned configuration of
$\TBs$ and the known data $D$.  In our discrete example, we compute
\begin{equation} \label{eq:enijk}
{\rm E}_{p(\bx|D,\sTBs,\hBs)}(N_{ijk}) = 
  \sum_{l=1}^N p(\x_i^k,\pai^j|\by_l,\TBs,\hBs)
\end{equation}
where $\by_l$ is the possibly incomplete $l$th case in $D$.  When
$\X_i$ and all the variables in $\Pai$ are observed in case $\bx_l$,
the term for this case requires a trivial computation: it is either
zero or one.  Otherwise, we can use any Bayesian network inference
algorithm to evaluate the term.  This computation is called the {\em
expectation step} of the EM algorithm.

Next, we use the expected sufficient statistics as if they were actual
sufficient statistics from a complete random sample $D_c$.  If we are
doing an ML calculation, then we determine the configuration of $\TBs$
that maximize $p(D_c|\TBs,\hBs)$.  In our discrete example, we have
\[
\tijk = \frac{{\rm E}_{p(\bx|D,\sTBs,\hBs)}(N_{ijk})}{
  \sum_{k=1}^{r_i} {\rm E}_{p(\bx|D,\sTBs,\hBs)}(N_{ijk})}
\]
If we are doing a MAP calculation, then we determine the configuration
of $\TBs$ that maximizes $p(\TBs|D_c,\hBs)$.  In our discrete example,
we have\footnote{The MAP configuration $\tilde{\TBs}$ depends on the
coordinate system in which the parameter variables are expressed.  The
expression for the MAP configuration given here is obtained by the
following procedure.  First, we transform each variable set
$\Tij=(\theta_{ij2},\ldots,\theta_{ijr_i})$ to the new coordinate
system $\phi_{ij}=(\phi_{ij2},\ldots,\phi_{ijr_i})$, where $\phi_{ijk}
= \log (\theta_{ijk}/\theta_{ij1}), k=2,\ldots,r_i$.  This coordinate
system, which we denote by $\phi_s$, is sometimes referred to as the
{\em canonical} coordinate system for the multinomial distribution
(see, e.g., Bernardo and Smith, 1994, pp. 199--202\nocite{BS94}).
Next, we determine the configuration of $\phi_s$ that maximizes
$p(\phi_s|D_c,\hBs)$.  Finally, we transform this MAP configuration to
the original coordinate system.  Using the MAP configuration
corresponding to the coordinate system $\phi_s$ has several
advantages, which are discussed in Thiesson
(1995b)\nocite{Thiesson95em} and MacKay (1996)\nocite{MacKay96}.}
\[
\tijk = \frac{\Npijk+{\rm E}_{p(\bx|D,\sTBs,\hBs)}(N_{ijk})}{
  \sum_{k=1}^{r_i} (\Npijk+{\rm E}_{p(\bx|D,\sTBs,\hBs)}(N_{ijk}))}
\]
This assignment is called the {\em maximization step} of the EM
algorithm.  Dempster {\em et al.} (1977) showed that, under certain
regularity conditions, iteration of the expectation and maximization
steps will converge to a local maximum.  The EM algorithm is typically
applied when sufficient statistics exist (i.e., when local
distribution functions are in the exponential family), although
generalizations of the EM algroithm have been used for more
complicated local distributions (see, e.g., Saul et
al. 1996\nocite{SJJ96}).

\section{Learning Parameters and Structure} \label{sec:struct}

Now we consider the problem of learning about both the structure and
probabilities of a Bayesian network given data.  

Assuming we think structure can be improved, we must be uncertain
about the network structure that encodes the physical joint
probability distribution for $\bX$.  Following the Bayesian approach,
we encode this uncertainty by defining a (discrete) variable whose
states correspond to the possible network-structure hypotheses $\hBs$,
and assessing the probabilities $p(\hBs)$.  Then, given a random
sample $D$ from the physical probability distribution for $\bX$, we
compute the posterior distribution $p(\hBs|D)$ and the posterior
distributions $p(\TBs|D,\hBs)$, and use these distributions in turn to
compute expectations of interest.  For example, to predict the next
case after seeing $D$, we compute
\begin{equation} \label{eq:fexp}
p(\bx_{N+1}|D) = \sum_{\hBs} p(\hBs|D) 
  \int p(\bx_{N+1}|\TBs,\hBs) \ p(\TBs|D,\hBs) \ d\TBs
\end{equation}
In performing the sum, we assume that the network-structure hypotheses
are mutually exclusive.  We return to this point in \Sec{sec:ml}.

The computation of $p(\TBs|D,\hBs)$ is as we have described in the
previous two sections.  The computation of $p(\hBs|D)$ is also
straightforward, at least in principle.  From Bayes' theorem, we have
\begin{equation} \label{eq:bd1}
p(\hBs|D) =  p(\hBs) \ p(D|\hBs) / p(D)
\end{equation}
where $p(D)$ is a normalization constant that does not depend upon
structure.  Thus, to determine the posterior distribution for network
structures, we need to compute the marginal likelihood of the data
($p(D|\hBs)$) for each possible structure.

We discuss the computation of marginal likelihoods in detail in
\Sec{sec:ml}.  As an introduction, consider our example with
unrestricted multinomial distributions, parameter independence,
Dirichlet priors, and complete data.  As we have discussed, when
there are no missing data, each parameter vector $\Tij$ is updated
independently.  In effect, we have a separate multi-sided thumbtack
problem for every $i$ and $j$.  Consequently, the marginal likelihood
of the data is the just the product of the marginal likelihoods for
each $i$--$j$ pair (given by \Eq{eq:dir-ml}):
\begin{equation} \label{eq:bd}
p(D|\hBs) = \prod_{i=1}^n \prod_{j=1}^{q_i}
\frac{\Gamma(\alpha_{ij})}{\Gamma(\alpha_{ij}+N_{ij})} \cdot
\prod_{k=1}^{r_i}
\frac{\Gamma(\alpha_{ijk} + N_{ijk})}{\Gamma(\alpha_{ijk})}  
\end{equation}  
This formula was first derived by Cooper and Herskovits
(1992)\nocite{CH92}.

Unfortunately, the full Bayesian approach that we have described is
often impractical.  One important computation bottleneck is produced
by the average over models in \Eq{eq:fexp}.  If we consider
Bayesian-network models with $n$ variables, the number of possible
structure hypotheses is more than exponential in $n$.  Consequently,
in situations where the user can not exclude almost all of these
hypotheses, the approach is intractable.

Statisticians, who have been confronted by this problem for decades in
the context of other types of models, use two approaches to address
this problem: {\em model selection} and {\em selective model
averaging}.  The former approach is to select a ``good'' model (i.e.,
structure hypothesis) from among all possible models, and use it as if it
were the correct model.  The latter approach is to select a manageable
number of good models from among all possible models and pretend that
these models are exhaustive.  These related approaches raise several
important questions.  In particular, do these approaches yield
accurate results when applied to Bayesian-network structures?  If so,
how do we search for good models?  And how do we decide whether or not
a model is ``good''?

\trj{ The question of accuracy is difficult to answer in theory.
Nonetheless, several researchers have shown experimentally that the
selection of a single good hypothesis often yields accurate
predictions (Cooper and Herskovits 1992; Aliferis and Cooper 1994;
Heckerman {\em et al.}, 1995b)\nocite{CH92,AC94uai,HGC95ml} and that model
averaging using Monte-Carlo methods can sometimes be efficient and
yield even better predictions (Madigan {\em et al.}, 1996\nocite{MRVH96}).
These results are somewhat surprising, and are largely responsible for
the great deal of recent interest in learning with Bayesian networks.
In Sections~\ref{sec:metric} through \ref{sec:priors}, we consider
different definitions of what is means for a model to be ``good'', and
discuss the computations entailed by some of these definitions.  In
\Sec{sec:search}, we discuss model search.

}{ 

The questions of accuracy and search are difficult to answer in
theory.  Nonetheless, several researchers have shown experimentally
that the selection of a single good hypothesis using greedy search
often yields accurate predictions (Cooper and Herskovits 1992;
Aliferis and Cooper 1994; Heckerman {\em et al.}, 1995b; Spirtes and Meek,
1995; Chickering, 1996)\nocite{CH92,AC94uai,HGC95ml,SM95kdd,C96uai},
and that model averaging using Monte-Carlo methods can sometimes be
efficient and yield even better predictions (Madigan {\em et al.},
1996\nocite{MRVH96}).  These results are somewhat surprising, and are
largely responsible for the great deal of recent interest in learning
with Bayesian networks.  In Sections~\ref{sec:metric} through
\ref{sec:priors}, we consider different definitions of what is means
for a model to be ``good'', and discuss the computations entailed by
some of these definitions.

}

We note that model averaging and model selection lead to models that
generalize well to {\em new} data.  That is, these techniques help us
to avoid the overfitting of data.  As is suggested by \Eq{eq:fexp},
Bayesian methods for model averaging and model selection are efficient
in the sense that all cases in $D$ can be used to both smooth and
train the model.  As we shall see in the following two sections, this
advantage holds true for the Bayesian approach in general.

\section{Criteria for Model Selection} \label{sec:metric}

Most of the literature on learning with Bayesian networks is concerned
with model selection.  In these approaches, some {\em criterion} is
used to measure the degree to which a network structure (equivalence
class) fits the prior knowledge and data.  A search algorithm is then
used to find an equivalence class that receives a high score by this
criterion.  Selective model averaging is more complex, because it is
often advantageous to identify network structures that are
significantly different.  In many cases, a single criterion is
unlikely to identify such complementary network structures.  In this
section, we discuss criteria for the simpler problem of model
selection.  For a discussion of selective model averaging, see Madigan
and Raftery (1994)\nocite{MR94}.

\subsection{Relative Posterior Probability}

A criterion that is often used for model selection is the log of the
relative posterior probability $\log p(D,\hBs) = \log p(\hBs) + \log
p(D|\hBs)$.\footnote{An equivalent criterion that is often used is
$\log (p(\hBs|D)/p(\hBs_0|D)) = \log (p(\hBs)/p(\hBs_0)) + \log
(p(D|\hBs)/p(D|\hBs_0))$.  The ratio $p(D|\hBs)/p(D|\hBs_0)$ is known
as a {\em Bayes' factor}.}  The logarithm is used for numerical
convenience.  This criterion has two components: the log prior and the
log marginal likelihood.  In \Sec{sec:ml}, we examine the computation
of the log marginal likelihood.  In \Sec{sec:spri}, we discuss the
assessment of network-structure priors.  Note that our comments about
these terms are also relevant to the full Bayesian approach.

The log marginal likelihood has the following interesting
interpretation described by Dawid (1984)\nocite{Dawid84}.  From the
chain rule of probability, we have
\begin{equation} \label{eq:preq}
\log p(D|\hBs) = \sum_{l=1}^N \log p(\bx_l|\bx_1,\ldots,\bx_{l-1},\hBs)
\end{equation}
The term $p(\bx_l|\bx_1,\ldots,\bx_{l-1},\hBs)$ is the prediction for
$\bx_l$ made by model $\hBs$ after averaging over its parameters.  The
log of this term can be thought of as the utility or reward for this
prediction under the utility function $\log p(\bx)$.\footnote{This
utility function is known as a {\em proper scoring rule}, because its
use encourages people to assess their true probabilities.  For a
characterization of proper scoring rules and this rule in particular,
see Bernardo (1979)\nocite{Bernardo79}.}  Thus, a model with the
highest log marginal likelihood (or the highest posterior probability,
assuming equal priors on structure) is also a model that is the best
sequential predictor of the data $D$ under the log utility function.

\begin{sloppypar}
Dawid (1984) also notes the relationship between this criterion and
cross validation.  When using one form of cross validation, known as
{\em leave-one-out} cross validation, we first train a model on all
but one of the cases in the random sample---say, $V_l=\{\bx_1,\ldots,
\bx_{l-1},\bx_{l+1},\ldots,\bx_{N}\}$.  Then, we predict the omitted
case, and reward this prediction under some utility function.  Finally,
we repeat this procedure for every case in the random sample, and sum
the rewards for each prediction.  If the prediction is probabilistic
and the utility function is $\log p(\bx)$, we obtain the
cross-validation criterion
\begin{equation} \label{eq:cv}
{\rm CV}(\hBs,D) = \sum_{l=1}^N \log p(\bx_l|V_l,\hBs)
\end{equation}
which is similar to \Eq{eq:preq}.  One problem with this criterion is
that training and test cases are interchanged.  For example, when we
compute $p(\bx_1|V_1,\hBs)$ in \Eq{eq:cv}, we use $\bx_2$ for training
and $\bx_1$ for testing.  Whereas, when we compute
$p(\bx_2|V_2,\hBs)$, we use $\bx_1$ for training and $\bx_2$ for
testing.  Such interchanges can lead to the selection of a model that
over fits the data (Dawid, 1984)\nocite{Dawid84}.  Various approaches
for attenuating this problem have been described, but we see from
\Eq{eq:preq} that the log-marginal-likelihood criterion avoids the
problem altogether.  Namely, when using this criterion, we never
interchange training and test cases.
\end{sloppypar}

\subsection{Local Criteria}

\begin{figure} 
\begin{center} 
\leavevmode 
\includegraphics[width=2.5in]{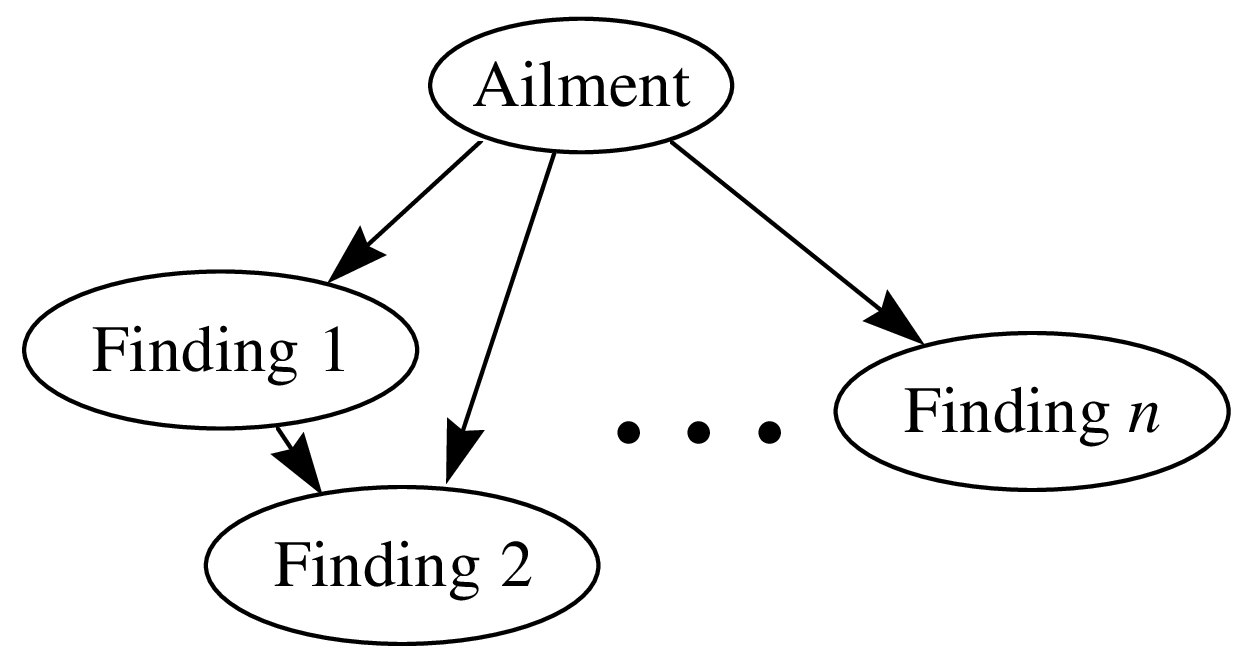}
\end{center}
\caption{A Bayesian-network structure for medical diagnosis.}
\label{fig:med-rxa}
\end{figure}

Consider the problem of diagnosing an ailment given the observation of
a set of findings.  Suppose that the set of ailments under
consideration are mutually exclusive and collectively exhaustive, so
that we may represent these ailments using a single variable $A$.  A
possible Bayesian network for this classification problem is shown in
\Fig{fig:med-rxa}.

The posterior-probability criterion is {\em global} in the sense that
it is equally sensitive to all possible dependencies.  In the
diagnosis problem, the posterior-probability criterion is just as
sensitive to dependencies among the finding variables as it is to
dependencies between ailment and findings.  Assuming that we observe
all (or perhaps all but a few) of the findings in $D$, a more
reasonable criterion would be {\em local} in the sense that it ignores
dependencies among findings and is sensitive only to the dependencies
among the ailment and findings.  This observation applies to all
classification and regression problems with complete data.

One such local criterion, suggested by Spiegelhalter et
al. (1993)\nocite{SDLC93}, is a variation on the sequential
log-marginal-likelihood criterion:
\begin{equation} \label{eq:locals}
{\rm LC}(\hBs,D) = \sum_{l=1}^N \log p(a_l|\bF_l,D_l,\hBs)
\end{equation}
where $a_l$ and $\bF_l$ denote the observation of the ailment $A$ and
findings $\bF$ in the $l$th case, respectively.  In other words, to
compute the $l$th term in the product, we train our model $\Bs$ with
the first $l-1$ cases, and then determine how well it predicts the
ailment given the findings in the $l$th case.  We can view this
criterion, like the log-marginal-likelihood, as a form of cross
validation where training and test cases are never interchanged.

The log utility function has interesting theoretical properties, but
it is sometimes inaccurate for real-world problems.  In general, an
appropriate reward or utility function will depend on the
decision-making problem or problems to which the probabilistic models
are applied.  Howard and Matheson (1983)\nocite{BlueBook} have
collected a series of articles describing how to construct utility
models for specific decision problems.  Once we construct such utility
models, we can use suitably modified forms of \Eq{eq:locals} for model
selection.

\section{Computation of the Marginal Likelihood} \label{sec:ml}

As mentioned, an often-used criterion for model selection is the log
relative posterior probability $\log p(D,\hBs) = \log p(\hBs) + \log
p(D|\hBs)$.  In this section, we discuss the computation of the 
second component of this criterion: the log marginal likelihood.

Given (1) local distribution functions in the exponential family, (2)
mutual independence of the parameters $\Thi$, (3) conjugate priors for
these parameters, and (4) complete data, the log marginal likelihood
can be computed efficiently and in closed form. \Eq{eq:bd} is an
example for unrestricted multinomial distributions.  Buntine (1994)
and Heckerman and Geiger (1996)\nocite{HG96tr} discuss the computation
for other local distribution functions.  Here, we concentrate on
approximations for incomplete data.

The Monte-Carlo and Gaussian approximations for learning about
parameters that we discussed in \Sec{sec:miss} are also useful for
computing the marginal likelihood given incomplete data.  One
Monte-Carlo approach, described by Chib (1995)\nocite{Chib95} and
Raftery (1996)\nocite{Raftery96b}, uses Bayes' theorem:
\begin{equation} \label{eq:chib}
p(D|\hBs) = \frac{p(\TBs|\hBs) \ p(D|\TBs,\hBs)}{p(\TBs|D,\hBs)}
\end{equation}
For any configuration of $\TBs$, the prior term in the numerator can
be evaluated directly.  In addition, the likelihood term in the
numerator can be computed using Bayesian-network inference.  Finally,
the posterior term in the denominator can be computed using Gibbs
sampling, as we described in \Sec{sec:gibbs}.  Other, more
sophisticated Monte-Carlo methods are described by DiCiccio et
al. (1995)\nocite{DKRW95}.

As we have discussed, Monte-Carlo methods are accurate but
computationally inefficient, especially for large data sets.  In
contrast, methods based on the Gaussian approximation are more
efficient, and can be as accurate as Monte-Carlo methods on large data
sets.

Recall that, for large amounts of data, $p(D|\TBs,\hBs) \cdot
p(\TBs|\hBs)$ can often be approximated as a multivariate-Gaussian
distribution.  Consequently,
\begin{equation} \label{eq:margl}
p(D|\hBs) = \int p(D|\TBs,\hBs) \ p(\TBs|\hBs) \ d\TBs
\end{equation}
can be evaluated in closed form.  In particular, substituting
\Eq{eq:post-gauss} into \Eq{eq:margl}, integrating, and taking
the logarithm of the result, we obtain the approximation:
\begin{equation} \label{eq:laplace}
\log p(D|\hBs) \approx \log p(D|\tilde{\TBs},\hBs) \ 
  + \log p(\tilde{\TBs}|\hBs) \ 
  + \frac{d}{2} \log (2\pi)
  - \frac{1}{2} \log |A|    
\end{equation}
where $d$ is the dimension of $g(\TBs)$.  For a Bayesian network with
unrestricted multinomial distributions, this dimension is typically
given by $\sum_{i=1}^n q_i(r_i-1)$.  Sometimes, when there are
hidden variables, this dimension is lower.  See Geiger et
al. (1996)\nocite{GHM96uai} for a discussion of this point.

This approximation technique for integration is known as {\em
Laplace's method}, and we refer to \Eq{eq:laplace} as the {\em Laplace
approximation}.  Kass {\em et al.} (1988) have shown that, under certain
regularity conditions, relative errors in this approximation are
$O(1/N)$, where $N$ is the number of cases in $D$.  Thus, the Laplace
approximation can be extremely accurate.  For more detailed
discussions of this approximation, see---for example---Kass et
al. (1988) and Kass and Raftery (1995)\nocite{KTK88,KR95}.

\erice{\begin{sloppypar}}{}
Although Laplace's approximation is efficient relative to Monte-Carlo
approaches, the computation of $|A|$ is nevertheless intensive for
large-dimension models.  One simplification is to approximate $|A|$
using only the diagonal elements of the Hessian $A$.  Although in so
doing, we incorrectly impose independencies among the parameters,
researchers have shown that the approximation can be accurate in some
circumstances (see, e.g., Becker and Le Cun, 1989\nocite{BL89}, and
Chickering and Heckerman, 1996\nocite{CH96tr}).  Another efficient
variant of Laplace's approximation is described by Cheeseman and Stutz
(1995)\nocite{CS95}, who use the approximation in the AutoClass
program for data clustering (see also Chickering and Heckerman,
1996\nocite{CH96tr}.)
\erice{\end{sloppypar}}{}

We obtain a very efficient (but less accurate) approximation by
retaining only those terms in \Eq{eq:laplace} that increase with $N$:
$\log p(D|\tilde{\TBs},\hBs)$, which increases linearly with $N$, and
$\log |A|$, which increases as $d \log N$.  Also, for large $N$,
$\tilde{\TBs}$ can be approximated by the ML configuration of $\TBs$.
Thus, we obtain
\begin{equation} \label{eq:bic}
\log p(D|\hBs) \approx \log p(D|\hat{\TBs},\hBs) \ 
  - \frac{d}{2} \log N
\end{equation}
This approximation is called the {\em Bayesian information criterion}
(BIC), and was first derived by Schwarz (1978).\nocite{Schwarz78}

The BIC approximation is interesting in several respects.  First, it
does not depend on the prior.  Consequently, we can use the
approximation without assessing a prior.\footnote{One of the technical
assumptions used to derive this approximation is that the prior is
non-zero around $\hat{\TBs}$.}  Second, the approximation is quite
intuitive.  Namely, it contains a term measuring how well the
parameterized model predicts the data ($\log p(D|\hat{\TBs},\hBs)$)
and a term that punishes the complexity of the model ($d/2 \ log N$).
Third, the BIC approximation is exactly minus the Minimum Description
Length (MDL) criterion described by Rissanen
(1987).\nocite{Rissanen87} Thus, recalling the discussion in
\Sec{sec:ml}, we see that the marginal likelihood provides a
connection between cross validation and MDL.

\section{Priors} \label{sec:priors}

To compute the relative posterior probability of a network structure,
we must assess the structure prior $p(\hBs)$ and the parameter priors
$p(\TBs|\hBs)$ (unless we are using large-sample approximations such
as BIC/MDL).  The parameter priors $p(\TBs|\hBs)$ are also required
for the alternative scoring functions discussed in
\Sec{sec:metric}.  Unfortunately, when many network structures are possible,
these assessments will be intractable.  Nonetheless, under certain
assumptions, we can derive the structure and parameter priors for many
network structures from a manageable number of direct assessments.
Several authors have discussed such assumptions and corresponding
methods for deriving priors (Cooper and Herskovits, 1991, 1992;
Buntine, 1991; Spiegelhalter {\em et al.}, 1993; Heckerman {\em et al.}, 1995b;
Heckerman and Geiger, 1996).\nocite{CH91tr,Buntine91,SDLC93,HGC95ml,HG96tr}
In this section, we examine some of these approaches.

\subsection{Priors on Network Parameters} \label{sec:ppri}

First, let us consider the assessment of priors for the model
parameters.  We consider the approach of Heckerman {\em et al.}
(1995b)\nocite{HGC95ml} who address the case where the local
distribution functions are unrestricted multinomial distributions and
the assumption of parameter independence holds.

Their approach is based on two key concepts: independence equivalence
and distribution equivalence.  We say that two Bayesian-network
structures for $\U$ are {\em independence equivalent} if they
represent the same set of conditional-independence assertions for $\U$
(Verma and Pearl, 1990)\nocite{Verma90}.  For example, given
$\bX=\{X,Y,Z\}$, the network structures $X
\rightarrow Y \rightarrow Z$, $X \leftarrow Y \rightarrow Z$, and $X
\leftarrow Y \leftarrow Z$ represent only the independence assertion that
$X$ and $Z$ are conditionally independent given $Y$.  Consequently, these
network structures are equivalent.  As another example, a {\em
complete network structure} is one that has no missing edge---that
is, it encodes no assertion of conditional independence.  When $\bX$
contains $n$ variables, there are $n!$ possible complete network
structures: one network structure for each possible ordering of the
variables.  All complete network structures for $p(\bx)$ are independence
equivalent.  In general, two network structures are independence
equivalent if and only if they have the same structure ignoring arc
directions and the same v-structures (Verma and Pearl,
1990).\nocite{Verma90} A {\em v-structure} is an ordered tuple
$(X,Y,Z)$ such that there is an arc from $X$ to $Y$ and from $Z$ to
$Y$, but no arc between $X$ and $Z$.

The concept of distribution equivalence is closely related to that of
independence equivalence.  Suppose that all Bayesian networks for $\U$
under consideration have local distribution functions in the family
${\cal F}$.  This is not a restriction, per se, because ${\cal F}$ can
be a large family.  We say that two Bayesian-network structures
$\Bsone$ and $\Bstwo$ for $\U$ are {\em distribution equivalent with
respect to (wrt) ${\cal F}$} if they represent the same joint
probability distributions for $\U$---that is, if, for every $\TBsone$,
there exists a $\TBstwo$ such that
$p(\bx|\TBsone,\hBsone)=p(\bx|\TBstwo,\hBstwo)$, and vice versa.

Distribution equivalence wrt some ${\cal F}$ implies independence
equivalence, but the converse does not hold.  For example, when ${\cal
F}$ is the family of generalized linear-regression models, the
complete network structures for $n\geq 3$ variables do not represent
the same sets of distributions.  Nonetheless, there are families
${\cal F}$---for example, unrestricted multinomial distributions and
linear-regression models with Gaussian noise---where independence
equivalence implies distribution equivalence wrt ${\cal F}$ (Heckerman
and Geiger, 1996)\nocite{HG96tr}.

The notion of distribution equivalence is important, because if two
network structures $\Bsone$ and $\Bstwo$ are distribution equivalent
wrt to a given ${\cal F}$, then the hypotheses associated with these
two structures are identical---that is, $\hBsone=\hBstwo$.  Thus, for
example, if $\Bsone$ and $\Bstwo$ are distribution equivalent, then
their probabilities must be equal in any state of information.
Heckerman {\em et al.} (1995b) call this property {\em hypothesis
equivalence}.

In light of this property, we should associate each hypothesis with an
equivalence class of structures rather than a single network
structure, and our methods for learning network structure should
actually be interpreted as methods for learning equivalence classes of
network structures (although, for the sake of brevity, we often blur
this distinction).  Thus, for example, the sum over network-structure
hypotheses in \Eq{eq:fexp} should be replaced with a sum over
equivalence-class hypotheses.  An efficient algorithm for identifying
the equivalence class of a given network structure can be found in
Chickering (1995)\nocite{C95uai}.

\erice{\begin{sloppypar}}{}
We note that hypothesis equivalence holds provided we interpret
Bayesian-network structure simply as a representation of conditional
independence.  Nonetheless, stronger definitions of Bayesian networks
exist where arcs have a causal interpretation (see \Sec{sec:cause}).
Heckerman {\em et al.} (1995b) and Heckerman (1995)\nocite{H95uai} argue
that, although it is unreasonable to assume hypothesis equivalence
when working with causal Bayesian networks, it is often reasonable to
adopt a weaker assumption of {\em likelihood equivalence,} which says
that the observational data can not help to discriminate two
indepence equivalent network structures.
\erice{\end{sloppypar}}{}

Now let us return to the main issue of this section: the derivation of
priors from a manageable number of assessments.  Geiger and Heckerman
(1995) show that the assumptions of parameter independence and
likelihood equivalence imply that the parameters for any {\em
complete} network structure $\Bsc$ must have a Dirichlet distribution
with constraints on the hyperparameters given by
\begin{equation} \label{eq:npijk}
\alpha_{ijk} = \alpha \ p(\x_i^k,\pai^j|\hBsc)
\end{equation}
where $\alpha$ is the user's equivalent sample size,\footnote{Recall
the method of equivalent samples for assessing beta and Dirichlet
distributions discussed in \Sec{sec:bayes}.}, and
$p(\x_i^k,\pai^j|\hBsc)$ is computed from the user's joint probability
distribution $p(\bx|\hBsc)$.\nocite{GH95tr-dir} This result is rather
remarkable, as the two assumptions leading to the constrained
Dirichlet solution are qualitative.

To determine the priors for parameters of {\em incomplete} network
structures, Heckerman {\em et al.} (1995b) use the assumption of {\em
parameter modularity,} which says that if $\X_i$ has the same parents
in network structures $\Bsone$ and $\Bstwo$, then
\[
p(\Tij|\hBsone) = p(\Tij|\hBstwo)
\]
for $j=1,\ldots,q_i$.  They call this property parameter modularity,
because it says that the distributions for parameters $\Tij$ depend
only on the structure of the network that is local to variable
$\X_i$---namely, $\X_i$ and its parents.  

\erice{\begin{sloppypar}}{}
Given the assumptions of parameter modularity and parameter
independence,\footnote{This construction procedure also assumes that
every structure has a non-zero prior probability.}  it is a simple
matter to construct priors for the parameters of an arbitrary network
structure given the priors on complete network structures.  In
particular, given parameter independence, we construct the priors for
the parameters of each node separately.  Furthermore, if node $X_i$
has parents $\Pai$ in the given network structure, we identify a
complete network structure where $X_i$ has these parents, and use
\Eq{eq:npijk} and parameter modularity to determine the priors for
this node.  The result is that all terms $\Npijk$ for all network
structures are determined by \Eq{eq:npijk}.  Thus, from the
assessments $\alpha$ and $p(\bx|\hBsc)$, we can derive the parameter
priors for all possible network structures.  Combining \Eq{eq:npijk}
with \Eq{eq:bd}, we obtain a model-selection criterion that assigns
equal marginal likelihoods to independence equivalent network
structures.
\erice{\end{sloppypar}}{}

We can assess $p(\bx|\hBsc)$ by constructing a Bayesian network,
called a {\em prior network}, that encodes this joint distribution.
Heckerman {\em et al.} (1995b) discuss the construction of this network.

\subsection{Priors on Structures} \label{sec:spri}

Now, let us consider the assessment of priors on network-structure
hypotheses.  Note that the alternative criteria described in
\Sec{sec:metric} can incorporate prior biases on network-structure
hypotheses.  Methods similar to those discussed in this section can be
used to assess such biases.

The simplest approach for assigning priors to network-structure
hypotheses is to assume that every hypothesis is equally likely.  Of
course, this assumption is typically inaccurate and used only for the
sake of convenience.  A simple refinement of this approach is to ask
the user to exclude various hypotheses (perhaps based on judgments of
of cause and effect), and then impose a uniform prior on the remaining
hypotheses.  We illustrate this approach in \Sec{sec:eg}.

Buntine (1991) describes a set of assumptions that leads to a richer
yet efficient approach for assigning priors.  The first assumption is
that the variables can be ordered (e.g., through a knowledge of time
precedence).  The second assumption is that the presence or absence of
possible arcs are mutually independent.  Given these assumptions,
$n(n-1)/2$ probability assessments (one for each possible arc in an
ordering) determines the prior probability of every possible
network-structure hypothesis.  One extension to this approach is to
allow for multiple possible orderings.  One simplification is to
assume that the probability that an arc is absent or present is
independent of the specific arc in question.  In this case, only one
probability assessment is required.

An alternative approach, described by Heckerman {\em et al.} (1995b) uses a
prior network.  The basic idea is to penalize the prior probability of
any structure according to some measure of deviation between that
structure and the prior network.  Heckerman {\em et al.} (1995b) suggest one
reasonable measure of deviation.

Madigan {\em et al.} (1995)\nocite{MGR95} give yet another approach that
makes use of imaginary data from a domain expert.  In their approach,
a computer program helps the user create a hypothetical set of
complete data.  Then, using techniques such as those in
\Sec{sec:struct}, they compute the posterior probabilities of
network-structure hypotheses given this data, assuming the prior
probabilities of hypotheses are uniform.  Finally, they use these
posterior probabilities as priors for the analysis of the real data.

\trj{
\section{Search Methods} \label{sec:search}

In this section, we examine search methods for identifying network
structures with high scores by some criterion.  Consider the problem
of finding the best network from the set of all networks in which each
node has no more than $k$ parents.  Unfortunately, the problem for
$k>1$ is NP-hard even when we use the restrictive prior given by
\Eq{eq:npijk} (Chickering {\em et al.} 1995)\nocite{CGH95aistats}.  Thus,
researchers have used heuristic search algorithms, including greedy
search, greedy search with restarts, best-first search, and
Monte-Carlo methods.

One consolation is that these search methods can be made more
efficient when the model-selection criterion is separable.  Given a
network structure for domain $\U$, we say that a criterion for that
structure is {\em separable} if it can be written as a product of
variable-specific criteria:
\begin{equation} \label{eq:score-loc} 
{\rm C}(\hBs,D) = \prod_{i=1}^n \ c(\X_i,\Pai,D_i)
\end{equation} 
where $D_i$ is the data restricted to the variables $\X_i$ and
$\Pai$.  An example of a separable criterion is the BD criterion
(\Eqs{eq:bd1} and \ref{eq:bd}) used in conjunction with any of the
methods for assessing structure priors described in
\Sec{sec:priors}.

Most of the commonly used search methods for Bayesian networks make
successive arc changes to the network, and employ the property of
separability to evaluate the merit of each change.  The possible
changes that can be made are easy to identify.  For any pair of
variables, if there is an arc connecting them, then this arc can
either be reversed or removed.  If there is no arc connecting them,
then an arc can be added in either direction.  All changes are subject
to the constraint that the resulting network contains no directed
cycles.  We use $E$ to denote the set of eligible changes to a graph,
and $\Delta(e)$ to denote the change in log score of the network
resulting from the modification $e
\in E$.  Given a separable criterion, if an arc to $\X_i$ is
added or deleted, only $c(\X_i,\Pai,D_i)$ need be evaluated to
determine $\Delta(e)$.  If an arc between $\X_i$ and $\X_j$ is
reversed, then only $c(\X_i,\Pai,D_i)$ and $c(\X_j,\Pi_j,D_j)$ need be
evaluated.
 
One simple heuristic search algorithm is greedy search.  First, we
choose a network structure.  Then, we evaluate $\Delta(e)$ for all $e
\in E$, and make the change $e$ for which $\Delta(e)$ is a maximum,
provided it is positive.  We terminate search when there is no $e$
with a positive value for $\Delta(e)$.  When the criterion is
separable, we can avoid recomputing all terms $\Delta(e)$ after every
change.  In particular, if neither $\X_i$, $\X_j$, nor their parents
are changed, then $\Delta(e)$ remains unchanged for all changes $e$
involving these nodes as long as the resulting network is acyclic.
Candidates for the initial graph include the empty graph, a random
graph, and a prior network.

A potential problem with any local-search method is getting stuck at a
local maximum.  One method for escaping local maxima is greedy search
with random restarts.  In this approach, we apply greedy search until
we hit a local maximum.  Then, we randomly perturb the network
structure, and repeat the process for some manageable number of
iterations.

Another method for escaping local maxima is simulated annealing.  In
this approach, we initialize the system at some temperature $T_0$.
Then, we pick some eligible change $e$ at random, and evaluate the
expression $p = \exp(\Delta(e)/T_0)$.  If $p > 1$, then we make the
change $e$; otherwise, we make the change with probability $p$.  We
repeat this selection and evaluation process $\alpha$ times or until
we make $\beta$ changes.  If we make no changes in $\alpha$
repetitions, then we stop searching.  Otherwise, we lower the
temperature by multiplying the current temperature $T_0$ by a decay
factor $0 < \gamma < 1$, and continue the search process.  We stop
searching if we have lowered the temperature more than $\delta$
times.  Thus, this algorithm is controlled by five parameters: $T_0,
\alpha, \beta, \gamma$ and $\delta$.  To initialize this algorithm, we can
start with the empty graph, and make $T_0$ large enough so that almost
every eligible change is made, thus creating a random graph.
Alternatively, we may start with a lower temperature, and use one of
the initialization methods described for local search.

Another method for escaping local maxima is best-first search (e.g.,
Korf, 1993)\nocite{Korf93}.  In this approach, the space of all
network structures is searched systematically using a heuristic
measure that determines the next best structure to examine.
Chickering (1996b)\nocite{Chickering96} has shown that, for a fixed
amount of computation time, greedy search with random restarts
produces better models than does best-first search.

One important consideration for any search algorithm is the search
space.  The methods that we have described search through the space of
Bayesian-network structures.  Nonetheless, when the assumption of
hypothesis equivalence holds, one can search through the space of
network-structure equivalence classes.  One benefit of the latter
approach is that the search space is smaller.  One drawback of the
latter approach is that it takes longer to move from one element in
the search space to another.  Work by Spirtes and Meek
(1995)\nocite{SM95kdd} and Chickering (1996)\nocite{C96uai} confirm
these observations experimentally.  Unfortunately, no comparisons are
yet available that determine whether the benefits of
equivalence-class search outweigh the costs.

}{}

\trj{\section{A Simple Example}}{\subsection{A Simple Example}}
\label{sec:eg}

Before we move on to other issues, let us step back and look at our
overall approach.  In a nutshell, we can construct both structure and
parameter priors by constructing a Bayesian network (the prior
network) along with additional assessments such as an equivalent
sample size and causal constraints.  We then use either Bayesian model
selection, selective model averaging, or full model averaging to
obtain one or more networks for prediction and/or explanation.  In
effect, we have a procedure for using data to improve the structure
and probabilities of an initial Bayesian network.

Here, we present \trj{two artificial examples}{a simple artificial
example} to illustrate this process.  Consider again the problem of
fraud detection from \Sec{sec:bn}.  Suppose we are given the data set
$D$ in \Tab{tab:db}, and we want to predict the next case---that is,
compute $p(\bx_{N+1}|D)$.  Let us assert that only two
network-structure hypotheses have appreciable probability: the
hypothesis corresponding to the network structure in \Fig{fig:bn}
($S_1$), and the hypothesis corresponding to the same structure with
an arc added from \node{Age} to \node{Gas} ($S_2$).  Furthermore, let
us assert that these two hypotheses are equally likely---that is,
$p(S^h_1)=p(S^h_2)=0.5$.  In addition, let us use the parameter priors
given by \Eq{eq:npijk}, where $\alpha=10$ and $p(\bx|\hBsc)$ is given
by the prior network in \Fig{fig:bn}.  Using Equations~\ref{eq:bd1}
and \ref{eq:bd}, we obtain $p(S^h_1|D)=0.26$ and $p(S^h_2|D)=0.74$.
Because we have only two models to consider, we can model average
according to \Eq{eq:fexp}:
\[
p(\bx_{N+1}|D) = 0.26 \ p(\bx_{N+1}|D,S^h_1) + 0.74 \ p(\bx_{N+1}|D,S^h_2) 
\]
where $p(\bx_{N+1}|D,\hBs)$ is given by \Eq{eq:nextc}.  (We don't
display these probability distributions\trj{.}{ for lack of space.})
If we had to choose one model, we would choose $S_2$, assuming the
posterior-probability criterion is appropriate.  Note that the data
favors the presence of the arc from \node{Age} to \node{Gas} by a
factor of three.  This is not surprising, because in the two cases in
the data set where fraud is absent and gas was purchased recently, the
card holder was less than 30 years old.

\begin{table} \label{tab:db}
\caption{An imagined data set for the fraud problem.}
\trsj{\begin{center}}{\begin{center}}{}{\begin{center}}
\begin{tabular}{|c|ccccc|}
\hline
Case & Fraud & Gas & Jewelry & Age    & Sex  \\ 
\hline
   1 & no    & no  & no      & 30-50  & female \\
   2 & no    & no  & no      & 30-50  & male \\
   3 & yes   & yes & yes     & $>$50    & male \\
   4 & no    & no  & no      & 30-50  & male \\
   5 & no    & yes & no      & $<$30    & female \\
   6 & no    & no  & no      & $<$30    & female \\
   7 & no    & no  & no      & $>$50    & male \\
   8 & no    & no  & yes     & 30-50  & female \\
   9 & no    & yes & no      & $<$30    & male \\
  10 & no    & no  & no      & $<$30    & female \\
\hline
\end{tabular}
\trsj{\end{center}}{\end{center}}{}{\end{center}}
\end{table} 

\trj{
An application of model selection, described by Spirtes and Meek
(1995), is illustrated in \Fig{fig:alarm}. \Fig{fig:alarm}a is a
hand-constructed Bayesian network for the domain of ICU ventilator
management, called the Alarm network (Beinlich et~al.,
1989)\nocite{Beinlich89}.  \Fig{fig:alarm}c is a random sample from
the Alarm network of size 10,000.  \Fig{fig:alarm}b is a simple prior
network for the domain.  This network encodes mutual independence
among the variables, and (not shown) uniform probability distributions
for each variable.

\Fig{fig:alarm}d shows the most likely network structure found
by a two-pass greedy search in equivalence-class space.  In the first
pass, arcs were added until the model score did not improve.  In the
second pass, arcs were deleted until the model score did not improve.
Structure priors were uniform; and parameter priors were computed from
the prior network using \Eq{eq:npijk} with $\alpha=10$.

The network structure learned from this procedure differs from the
true network structure only by a single arc deletion.  In effect, we
have used the data to improve dramatically the original model of the
user.  

\begin{figure}
\vspace{.5in}
\begin{center} 
\leavevmode 
\includegraphics[width=6.0in]{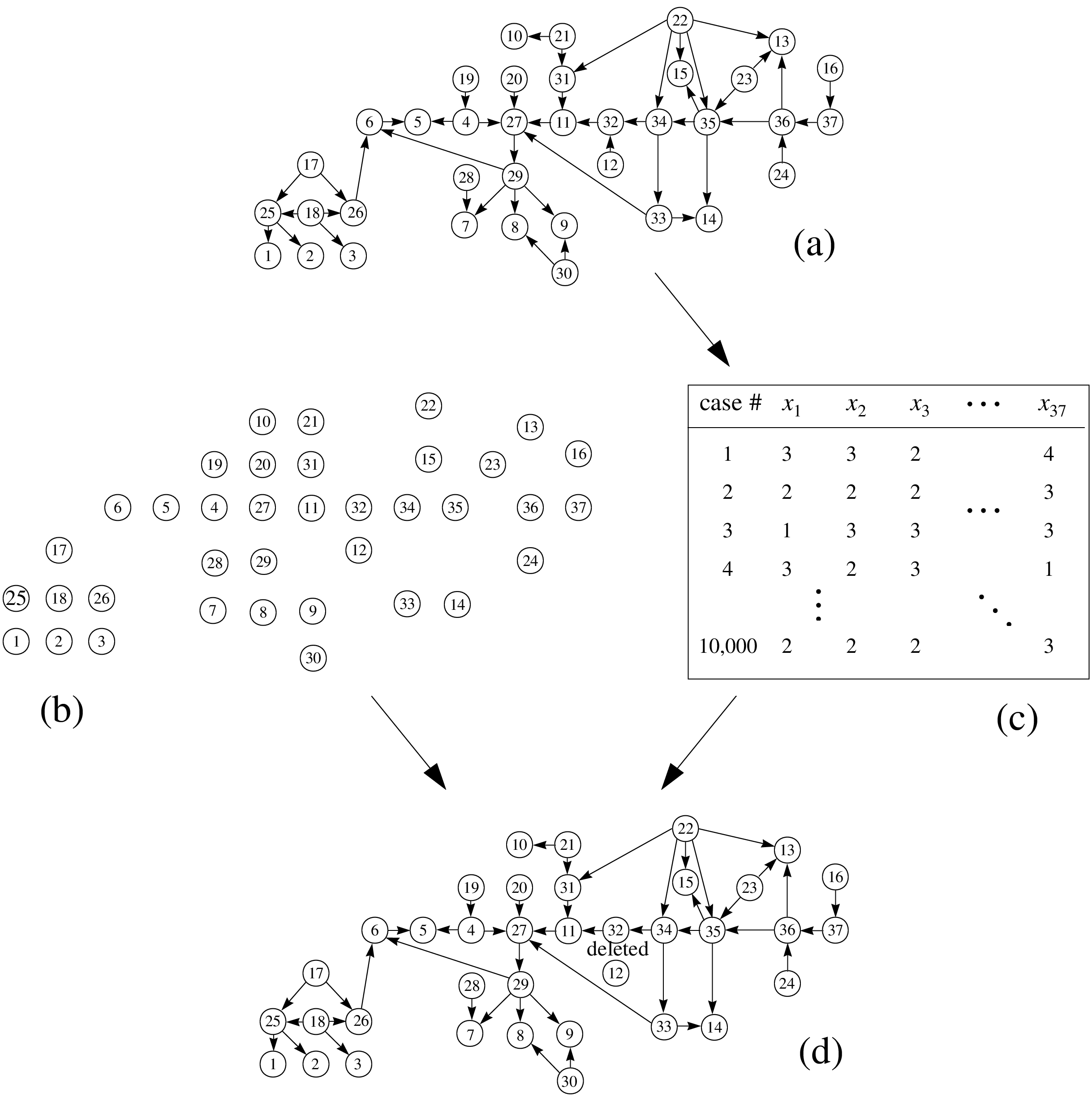}
\end{center}
\caption{
(a) The Alarm network structure.  (b) A prior network encoding a
user's beliefs about the Alarm domain.  (c) A random sample of size
10,000 generated from the Alarm network.  (d) The network learned from
the prior network and the random sample.  The only difference
between the learned and true structure is an arc deletion as noted
in (d).  Network probabilities are not shown.}
\label{fig:alarm}
\end{figure}

}{}

\section{Bayesian Networks for Supervised Learning} \label{sec:sup}

As we discussed in \Sec{sec:nvar}, the local distribution functions
$p(x_i|\pai,\Thi,\hBs)$ are essentially classification/regression
models.  Therefore, if we are doing supervised learning where the
explanatory (input) variables cause the outcome (target) variable and
data is complete, then the Bayesian-network and
classification/regression approaches are identical.

When data is complete but input/target variables do not have a simple
cause/effect relationship, tradeoffs emerge between the
Bayesian-network approach and other methods.  For example, consider
the classification problem in \Fig{fig:med-rxa}.  Here, the Bayesian
network encodes dependencies between findings and ailments as well as
among the findings, whereas another classification model such as a
decision tree encodes only the relationships between findings and
ailment.  Thus, the decision tree may produce more accurate
classifications, because it can encode the necessary relationships
with fewer parameters.  Nonetheless, the use of local criteria for
Bayesian-network model selection mitigates this advantage.
Furthermore, the Bayesian network provides a more natural
representation in which to encode prior knowledge, thus giving this
model a possible advantage for sufficiently small sample sizes.
Another argument, based on bias--variance analysis, suggests that
neither approach will dramatically outperform the other (Friedman,
1996).\nocite{Friedman96}

Singh and Provan (1995)\nocite{SP95} compare the classification
accuracy of Bayesian networks and decision trees using complete data
sets from the University of California, Irvine Repository of Machine
Learning data sets.  Specifically, they compare C4.5 with an algorithm
that learns the structure and probabilities of a Bayesian network
using a variation of the Bayesian methods we have described.  The
latter algorithm includes a model-selection phase that discards some
input variables.  They show that, overall, Bayesian networks and
decisions trees have about the same classification error.  These
results support the argument of Friedman (1996).

When the input variables cause the target variable and data is
incomplete, the dependencies between input variables becomes
important, as we discussed in the introduction.  Bayesian networks
provide a natural framework for learning about and encoding these
dependencies.  Unfortunately, no studies have been done comparing
these approaches with other methods for handling missing data.

\section{Bayesian Networks for Unsupervised Learning} \label{sec:unsup}

\begin{figure} 
\begin{center} 
\leavevmode 
\includegraphics[width=3in]{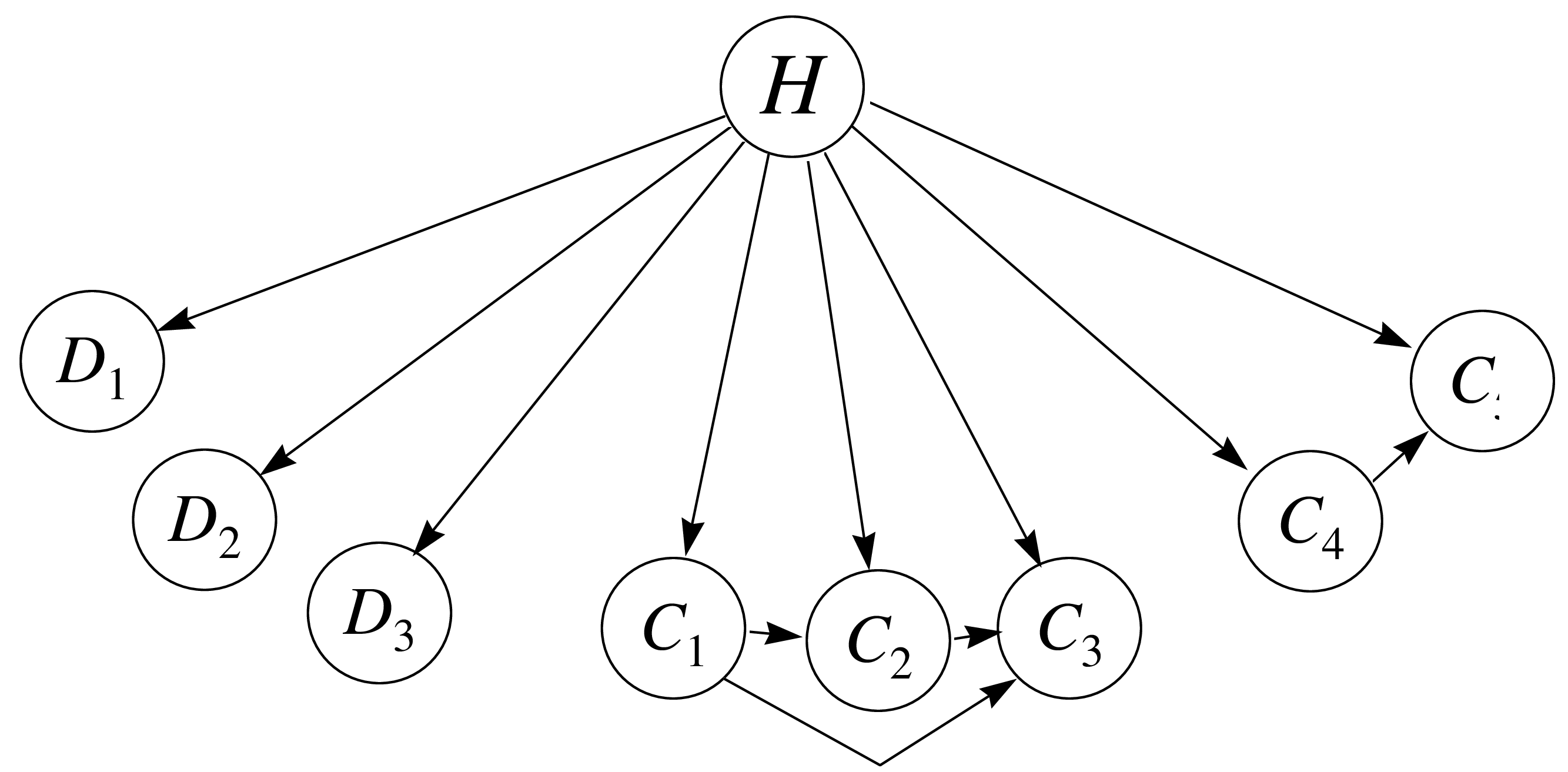}
\end{center}
\caption{A Bayesian-network structure for AutoClass.  The variable $H$ is
hidden.  Its possible states correspond to the underlying classes in the
data.}
\label{fig:ac}
\end{figure}

The techniques described in this paper can be used for unsupervised
learning.  A simple example is the AutoClass program of Cheeseman and
Stutz (1995)\nocite{CS95}, which performs data clustering.  The idea
behind AutoClass is that there is a single hidden (i.e., never
observed) variable that causes the observations.  This hidden variable
is discrete, and its possible states correspond to the underlying
classes in the data.  Thus, AutoClass can be described by a Bayesian
network such as the one in \Fig{fig:ac}.  For reasons of computational
efficiency, Cheeseman and Stutz (1995) assume that the discrete
variables (e.g., $D_1,D_2,D_3$ in the figure) and user-defined sets of
continuous variables (e.g., $\{C_1,C_2,C_3\}$ and $\{C_4,C_5\}$) are
mutually independent given $H$.  Given a data set $D$, AutoClass
searches over variants of this model (including the number of states
of the hidden variable) and selects a variant whose (approximate)
posterior probability is a local maximum.

AutoClass is an example where the user presupposes the existence of a
hidden variable.  In other situations, we may be unsure about the
presence of a hidden variable.  In such cases, we can score models
with and without hidden variables to reduce our uncertainty.  We
illustrate this approach on a real-world case study in \Sec{sec:app}.
Alternatively, we may have little idea about what hidden variables to
model.  The search algorithms of Spirtes {\em et al.} (1993) provide one
method for identifying possible hidden variables in such situations.
Martin and VanLehn (1995)\nocite{MvL95} suggest another method.

\begin{figure}
\begin{center} 
\leavevmode 
\includegraphics[width=4in]{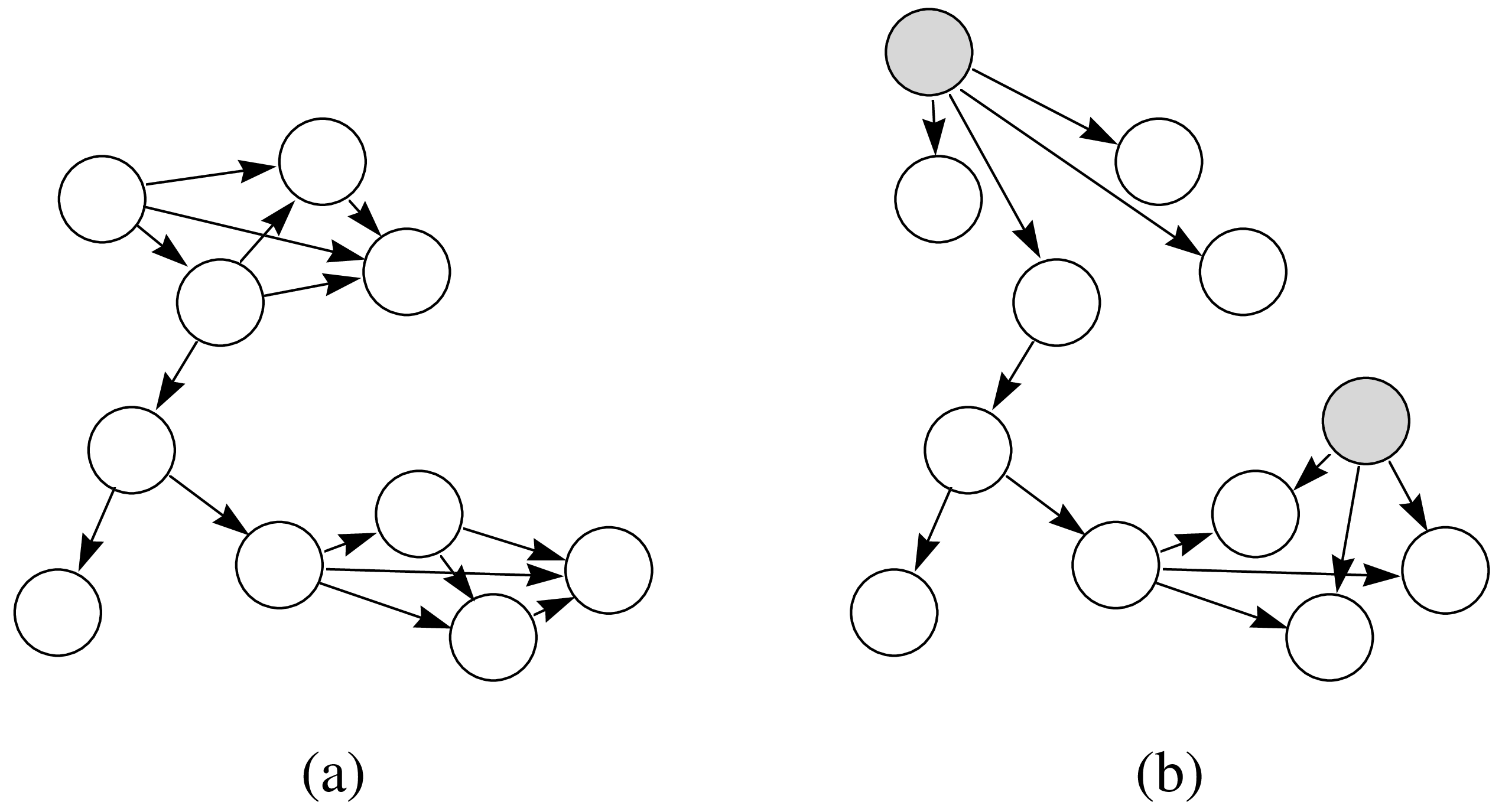}
\end{center}
\caption{
(a) A Bayesian-network structure for observed variables.  (b) A
Bayesian-network structure with hidden variables (shaded) suggested by
the network structure in (a).}
\label{fig:clusters}
\end{figure}

Their approach is based on the observation that if a set of variables
are mutually dependent, then a simple explanation is that these
variables have a single hidden common cause rendering them mutually
independent.  Thus, to identify possible hidden variables, we first
apply some learning technique to select a model containing no hidden
variables.  Then, we look for sets of mutually dependent variables in
this learned model.  For each such set of variables (and combinations
thereof), we create a new model containing a hidden variable that
renders that set of variables conditionally independent.  We then
score the new models, possibly finding one better than the original.
For example, the model in \Fig{fig:clusters}a has two sets of mutually
dependent variables.  \Fig{fig:clusters}b shows another model
containing hidden variables suggested by this model.

\section{Learning Causal Relationships} \label{sec:cause}

As we have mentioned, the causal semantics of a Bayesian network
provide a means by which we can learn causal relationships.  In this
section, we examine these semantics, and provide a basic discussion on
how causal relationships can be learned.
We note that these methods
are new and controversial.\footnote{There was
certainly controversy in 1996 when I wrote this tutorial.
Now in 2020, these ideas are much more accepted.  I leave this
comment here as a historical remark.}
For critical discussions on both sides of
the issue, see Spirtes {\em et al.} (1993)\nocite{SGS93}, Pearl
(1995)\nocite{Pearl95bm}, and Humphreys and Freedman
(1995)\nocite{HF96}.

For purposes of illustration, suppose we are marketing analysts who
want to know whether we should increase the placement of a magazine ad
to increase sales of a product.  If seeing the ad is a cause of
busying the product, then we want to show more of it.  More generally,
causal knowledge is at the heart of understanding what will happen
when we intervene.  For detailed discussions of the meaning of cause
and the close connections between causal knowledge and the consequences of
intervention, see Pearl (1995)\nocite{Pearl95bm} and Heckerman and
Shachter (1995)\nocite{HS95}.

How do we learn causal relationships from data?  One approach 
is to actually intervene, and note the consequences.  In the classic
embodiment of this approach, we perform a randomized experiment.  
In the ad example, we would (1) select a set of individuals at random, (2)
for each individual, show them the ad if and only if a coin flip comes
up heads, and (3) note any difference in buying behavior between the
two groups.  Unfortunately, randomized experiments can be expensive (as
in this example), unethical, or difficult to obtain compliance.

Over the last century, many researchers have developed methods for
inferring cause and effect.  Here, we consider a new approach that
makes use of a Bayesian network with causal semantics to infer causal
knowledge from observational data alone.  The basic idea is to {\em
  assume} a connection between causal and probabilistic relationships
in a directed network.  More precisely, we say that a directed acyclic
graph ${\cal C}$ is a {\em causal graph for variables $\bX$} if the
nodes in ${\cal C}$ are in a one-to-one correspondence with $\bX$, and
there is an arc from node $X$ to node $Y$ in ${\cal C}$ if and only if
$X$ is a direct cause of $Y$.  We now assume that if ${\cal C}$ is a
causal graph for $\bX$, then ${\cal C}$ is also a Bayesian-network
structure for the joint physical probability distribution of $\bX$.
This assumption, known as the {\em causal Markov condition}, is
considered to be quite weak ({\em i.e.,} quite reasonable) by many
philosophers (Spirtes {\em et al.}, 1993).\nocite{SGS93} We can infer
causal relationships from observational data by first inferring
probabilistic independence relationships as we have described earlier
in this tutorial, and then inferring causal relationships using this
assumption.\footnote{This approach also requires the assumption of
  {\em faithfulness}, which says that causal relationships do not
  accidentally produce probabilistic independence.  In our Bayesian
  approach to learning networks, however, faithfulness follows from
  our assumption that $p(\TBs|\hBs)$ is a probability density
  function.}  As we shall see, we can not always learn causal
relationships with this approach.  Whether we can, will depend on the
probabilistic relationships we find.

\begin{figure}
\begin{center} 
\leavevmode 
\includegraphics[width=4in]{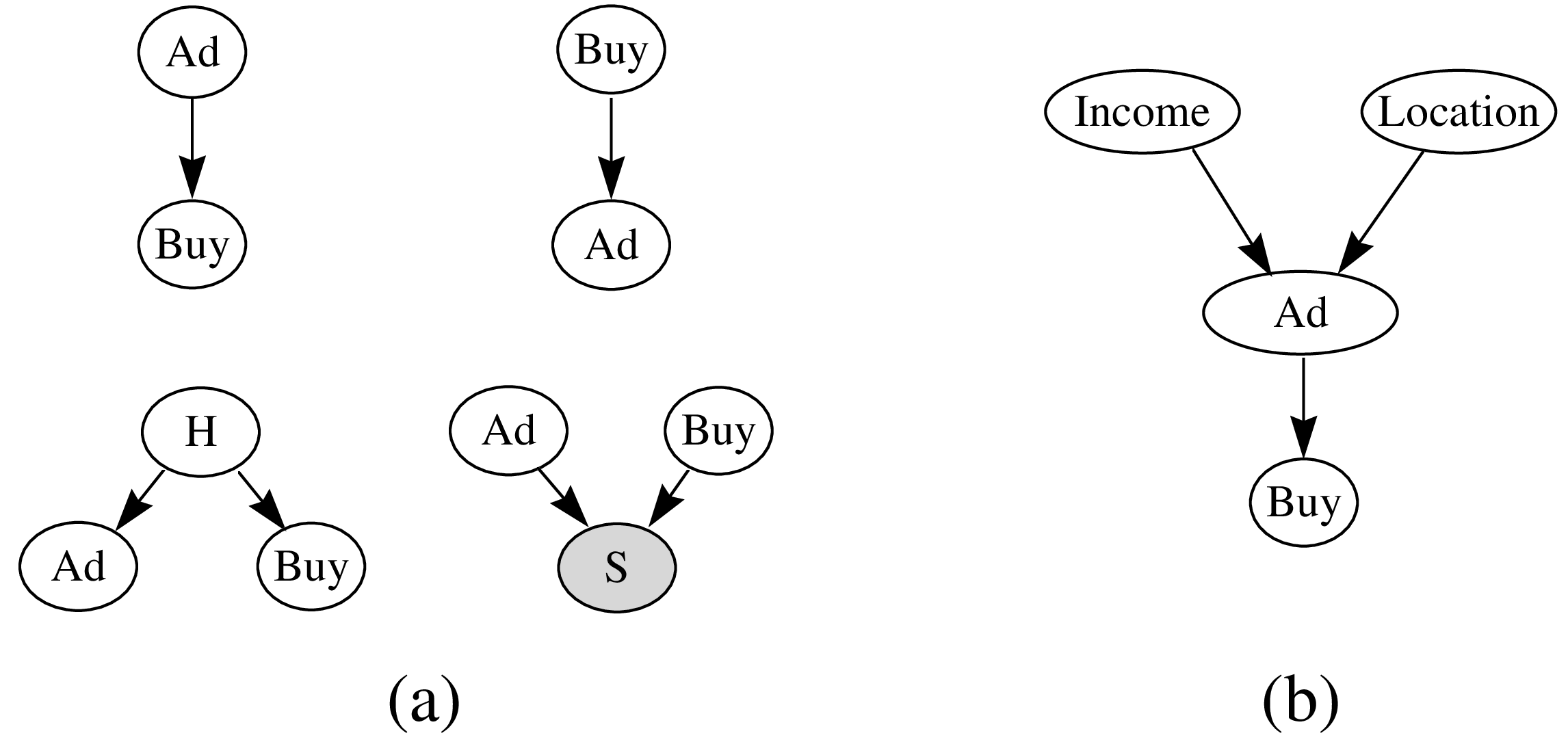}
\end{center}
\caption{
(a) Causal graphs showing four explanations for an observed dependence
between \node{Ad} and \node{Buy}.  The node $H$ corresponds to a
hidden common cause of \node{Ad} and \node{Buy}.  The shaded node
$S$ indicates that the case has been included in the data set. (b) A
Bayesian network for which \node{Ad} causes \node{Buy} is the only causal
explanation, given the causal Markov condition.}
\label{fig:cause}
\end{figure}

To illustrate this approach, let us return to the ad example.  Let
variables \node{Ad} ($A$) and \node{Buy} ($B$) represent whether or
not an individual has seen the advertisement and has purchased the
product, respectively.  Assuming these variables are dependent, there
are four simple causal explanations: (1) $A$ is a cause of $B$, (2)
$B$ is a cause of $A$, (3) there are one or more hidden common causes
of $A$ and $B$ (e.g., the person's gender), and (4) $A$ and $B$ are
causes for data selection.  This last explanation is an example of
{\em selection bias}.  Condition 4 would occur, for example, if our
data set failed to include instances where $A$ and $B$ are false.
These four causal explanations for the presence of the arcs are
illustrated in \Fig{fig:cause}a.  Of course, more complicated
explanations---such as the presence of a hidden common cause and
selection bias---are possible.

When we have observations only for $A$ and $B$, it is not possible to
distinguish among these various causal hypotheses.  Suppose, however,
that we observe two additional variables: \node{Income} ($I$) and
\node{Location} ($L$), which represent the income and geographic
location of the possible purchaser, respectively.  Furthermore,
suppose we learn (with high probability) the Bayesian network shown in
\Fig{fig:cause}b.  Given the causal Markov condition, the {\em only}
causal explanation for the conditional-independence and
conditional-dependence relationships encoded in this Bayesian network
is that \node{Ad} is a cause for \node{Buy}.  More precisely, none of
the other explanations described in the previous paragraph, or
combinations thereof, produce the probabilistic relationships encoded
in \Fig{fig:cause}b.

\section{A Case Study: College Plans} \label{sec:app}

Real-world applications of techniques that we have discussed can be
found in Madigan and Raftery (1994), Lauritzen et
al. (1994)\nocite{LTS94}, Singh and Provan (1995)\nocite{SP95}, and
Friedman and Goldszmidt (1996)\nocite{FG96}.  Here, we consider an
application that comes from a study by Sewell and Shah
(1968)\nocite{SS68}, who investigated factors that influence the
intention of high school students to attend college.  The data have
been analyzed by several groups of statisticians, including Whittaker
(1990)\nocite{Whittaker90} and Spirtes {\em et al.} (1993), all of whom have
used non-Bayesian techniques.

Sewell and Shah (1968) measured the following variables for 10,318
Wisconsin high school seniors: \node{Sex} (SEX): male, female;
\node{Socioeconomic Status} (SES): low, lower middle, upper middle,
high; \node{Intelligence Quotient} (IQ): low, lower middle, upper
middle, high; \node{Parental Encouragement} (PE): low, high; and
\node{College Plans} (CP): yes, no.  Our goal here is to
understand the (possibly causal) relationships among these variables.

The data are described by the sufficient statistics in \Tab{tab:cp}.
Each entry denotes the number of cases in which the five variables
take on some particular configuration.  The first entry corresponds to
the configuration SEX$=$male, SES$=$low, IQ$=$low, PE$=$low, and
$CP$=yes.  The remaining entries correspond to configurations obtained
by cycling through the states of each variable such that the last
variable (CP) varies most quickly.  Thus, for example, the upper
(lower) half of the table corresponds to male (female) students.

\begin{table} \label{tab:cp}
\caption{Sufficient statistics for the Sewall and Shah (1968) study.}
\trsj{\begin{center}}{\begin{center}}{}{\begin{center}}
\begin{tabular}{|cccccccccccccccc|}
\hline
  4&349& 13& 64&  9&207& 33& 72& 12&126& 38& 54& 10& 67& 49& 43 \\
  2&232& 27& 84&  7&201& 64& 95& 12&115& 93& 92& 17& 79&119& 59 \\
  8&166& 47& 91&  6&120& 74&110& 17& 92&148&100&  6& 42&198& 73 \\
  4& 48& 39& 57&  5& 47&123& 90&  9& 41&224& 65&  8& 17&414& 54 \\
   &   &   &   &   &   &   &   &   &   &   &   &   &   &   &    \\
  5&454&  9& 44&  5&312& 14& 47&  8&216& 20& 35& 13& 96& 28& 24 \\
 11&285& 29& 61& 19&236& 47& 88& 12&164& 62& 85& 15&113& 72& 50 \\
  7&163& 36& 72& 13&193& 75& 90& 12&174& 91&100& 20& 81&142& 77 \\
  6& 50& 36& 58&  5& 70&110& 76& 12& 48&230& 81& 13& 49&360& 98 \\
\hline
\end{tabular}
\trsj{\end{center}}{\end{center}}{}{\end{center}}
\noindent Reproduced by permission from the University of Chicago
Press.  \copyright 1968 by The University of Chicago.  All rights
reserved.
\end{table} 

\begin{figure}
\begin{center}
\leavevmode
\includegraphics[width=4.0in]{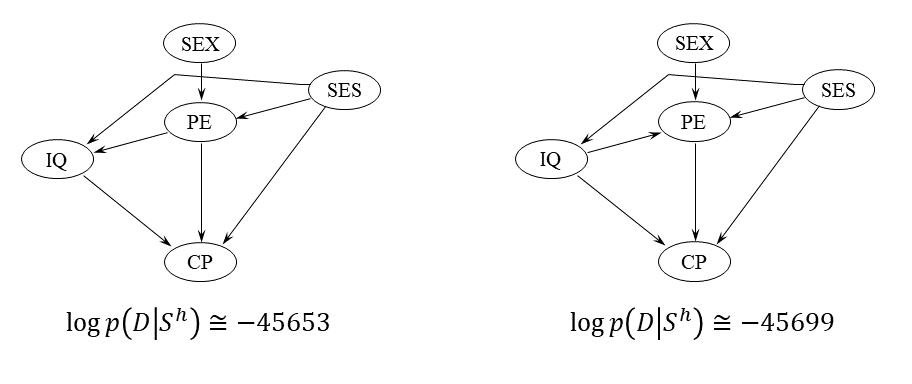}
\end{center}
\caption{The two network structures without hidden variables 
with the highest marginal likelihoods.}
\label{fig:cp1}
\end{figure}

As a first pass, we analyzed the data assuming there are no hidden
variables.  To generate priors for network parameters, we used the
method described in \Sec{sec:ppri} with an equivalent sample size of 5
and a prior network where $p(\bx|\hBsc)$ is uniform.  (The results
were not sensitive to the choice of parameter priors.  For example,
none of the results reported in this section changed qualitatively for
equivalent sample sizes ranging from 3 to 40.)  We considered all
possible structures except those where $SEX$ and/or $SES$ had parents,
and/or $CP$ had children.  Because the data set was complete, we used
\Eqs{eq:bd1} and \ref{eq:bd} to compute the marginal likelihoods.  The
two network structures with the highest marginal likelihoods are shown
in \Fig{fig:cp1}.

If we adopt the causal Markov assumption and also assume that there
are no hidden variables, then the arcs in both
graphs can be interpreted causally.  Some results are not
surprising---for example the causal influence of socioeconomic status
and IQ on college plans.  Other results are more interesting.  For
example, from either graph we conclude that sex influences college
plans only indirectly through parental influence.  Also, the two
graphs differ only by the orientation of the arc between PE and IQ.
Either causal relationship is plausible.  We note that the second most
likely graph was selected by Spirtes {\em et al.} (1993), who used a
non-Bayesian approach to infer the network.

The most suspicious result is the suggestion that socioeconomic status
has a direct influence on IQ.  To question this result, we considered
new network structures obtained from those in \Fig{fig:cp1} by
replacing this direct influence with a hidden variable pointing to
both $SES$ and $IQ$, and by inserting a hidden variable between $SES$
and $IQ$.  We also considered structures where the hidden variable
pointed to $SES$, $IQ$, and $PE$, and none, one, or both of the
connections $SES$---$PE$ and $PE$---$IQ$ were removed.  For each
structure, we varied the number of states of the hidden variable from
two to six.

\begin{figure}
\begin{center}
\leavevmode
\includegraphics[width=5.0in]{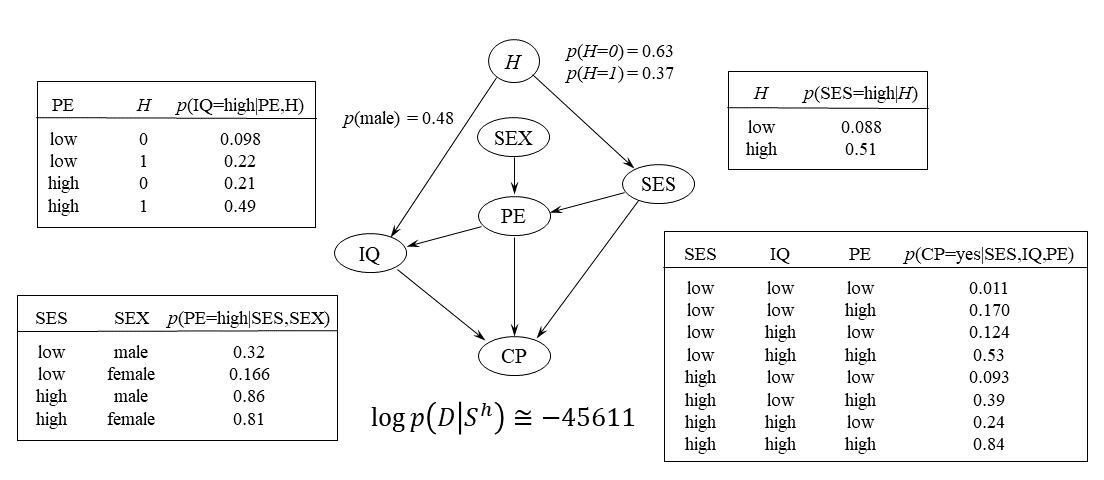}
\end{center}
\caption{One of the two hidden-variable network structures with the
  highest marginal likelihood.  The other model, which has the same
  marginal likelihood, has the arc from $H$ to SES reversed.
  Probabilities shown are MAP values.  Some probability distributions
  are omitted.}
\label{fig:cp2}
\end{figure}

We computed the marginal likelihoods of these structures using
the Cheeseman-Stutz (1995)\nocite{CS95} variant of the Laplace
approximation.  To find the MAP $\tilde{\TBs}$, we used the EM
algorithm, taking the largest local maximum from among 100 runs with
different random initializations of $\TBs$.  We then computed more
accurate marginal likelihoods for the best networks using annealed
importance sampling, a Monte-Carlo technique (Neal, 1998)\nocite{Neal98}.  
Among the
structures considered, there were two with equal marginal likelihoods,
much higher than the marginal likelihoods of the best structures
without hidden variables.  One is shown in \Fig{fig:cp2}.  The other
has the arc from $H$ to SES reversed.  In the model not shown in the
figure, SES is a cause of IQ as in the best model with no hidden
variables, and $H$ helps to capture the orderings of the states in SES
and IQ that are ignored in the multinomial model.  In the model shown
in the figure, $H$ again helps to capture the ordering of states, but
is also a hidden common cause of IQ and SES ({\em e.g.,}, parent
``quality.'').

In closing, we highlight that model averaging, selective or full, 
can be applied to Bayesian networks with causal semantics. Applying
selective model averaging to the two causal models we have
just discussed under the assumption that they are {\em a priori}
equally likely, the posterior probability
that ``$H$ causes $SES$'' would be 1/2. Furthermore, the posterior
probability of $SES$ given that we intervene on $H$ would
be the simple average of the probabilities under the two models.

\section{Pointers to Literature and Software} \label{sec:point}

Like all tutorials, this one is incomplete.  For those readers
interested in learning more about graphical models and methods for
learning them, we offer the following additional references and
pointers to software.  Buntine (1996) provides another guide to the
literature.\nocite{Buntine96}

Spirtes {\em et al.} (1993) and Pearl (1995) use methods based on
large-sample approximations to learn Bayesian networks.  In addition,
as we have discussed, they describe methods for learning causal
relationships from observational data.  

\erice{\begin{sloppypar}}{}
In addition to directed models, researchers have explored network
structures containing undirected edges as a knowledge representation.
These representations are discussed (e.g.) in Lauritzen
(1982)\nocite{Lauritzen82}, Verma and Pearl (1990)\nocite{Verma90},
Frydenberg (1990)\nocite{Frydenberg90}, Whittaker
(1990)\nocite{Whittaker90}, and Richardson (1997)\nocite{R97aistats}.
Bayesian methods for learning such models from data are described by
Dawid and Lauritzen (1993) and Buntine (1994).
\erice{\end{sloppypar}}{}

Finally, several research groups have developed software systems for
learning graphical models.  For example, Scheines {\em et al.} (1994) have
developed a software program called TETRAD II for learning about cause
and effect.  Badsberg (1992)\nocite{Badsberg92} and H{\o}jsgaard et
al. (1994)\nocite{HST94} have built systems that can learn with mixed
graphical models using a variety of criteria for model selection.
Thomas, Spiegelhalter, and Gilks (1992)\nocite{TSG92} have created a
system called BUGS that takes a learning problem specified as a
Bayesian network and compiles this problem into a Gibbs-sampler
computer program.

\section*{Acknowledgments} 
I thank Radford Neal and Max Chickering for scoring the Sewall and
Shah network structures with hidden variables, and Chris Meek for
bringing this data set to my attention.  I thank Max Chickering, Usama
Fayyad, Eric Horvitz, Chris Meek, Koos Rommelse, and Padhraic Smyth
for their comments on earlier versions of this manuscript.

\trsj{}{}{}{
\bigskip
\bigskip
\bigskip
\bigskip
\bigskip
\bigskip
\bigskip
\bigskip
\bigskip
\bigskip
\bigskip
\bigskip
\bigskip
\bigskip
\bigskip
\bigskip
\bigskip
\bigskip
\bigskip
\bigskip
\ \ \ 
}

\trj{
\section*{Notation} \label{sec:notation}

\begin{tabular}{rl}
$X, Y, Z, \ldots$  
                     & Variables or their corresponding nodes in a Bayesian \\
                     & network \\
$\bX, \bY, \bZ, \ldots$    
                     & Sets of variables or corresponding sets of nodes \\
$X=x$                & Variable $X$ is in state $x$ \\
$\bX=\bx$            & The set of variables $\bX$ is in configuration $\bx$ \\
$\bx, \by, \bz$
                     & Typically refer to a complete case, an incomplete \\
                     & case, and missing data in a case, respectively \\
$\bX \setminus \bY$  & The variables in $X$ that are not in $Y$ \\
$D$                  & A data set: a set of cases \\
$D_l$                & The first $l-1$ cases in $D$ \\
$p(\bx|\by)$ 
                     & The probability that $\bX=\bx$ given $\bY=\by$ \\
                     & (also used to describe a probability density, \\
                     & probability distribution, and probability density) \\
${\rm E}_{p(\cdot)}(x)$ 
                     & The expectation of $x$ with respect to $p(\cdot)$ \\
$\Bs$                & A Bayesian network structure 
                       (a directed acyclic graph) \\
$\Pai$               & The variable or node corresponding to the parents \\
                     & of node $X_i$ in a Bayesian network structure \\
$\pai$               & A configuration of the variables $\Pai$ \\
$r_i$                & The number of states of discrete variable $X_i$\\
$q_i$                & The number of configurations of $\Pai$\\
$\Bsc$               & A complete network structure \\
$\hBs$               & The hypothesis corresponding to network
                       structure $\Bs$\\
$\tijk$
                     & The multinomial parameter corresponding to the \\
                     & probability $p(X_i=x_i^k|\Pai=\pai^j)$ \\
$\Tij$               & $= (\theta_{ij2},\ldots,\theta_{ijr_i})$ \\
$\Th_i$              & $= (\Th_{i1},\ldots,\Th_{iq_i})$ \\
$\TBs$               & $= (\Th_1,\ldots,\Th_n)$ \\
$\alpha$             & An equivalent sample size \\
$\alpha_{ijk}$       & The Dirichlet hyperparameter corresponding to $\tijk$ \\
$\alpha_{ij}$        & $=\sum_{k=1}^{r_i} \alpha_{ijk}$ \\
$N_{ijk}$            & The number of cases in data set $D$ where
$X_i=x^k_i$ and $\Pai=\pai^j$ \\
$N_{ij}$             & $=\sum_{k=1}^{r_i} N_{ijk}$\\
\ & \ \\
\ & \ \\
\ & \ \\
\ & \ \\
\end{tabular}
}{}

\bibliographystyle{apalike} 

\bibliography{David}  

\section*{References}

Aliferis, C. and Cooper, G. (1994).
\newblock An evaluation of an algorithm for inductive learning of {Bayesian}
  belief networks using simulated data sets.
\newblock In {\em Proceedings of Tenth Conference on Uncertainty in Artificial
  Intelligence, {\rm Seattle, WA}}, pages 8--14. Morgan Kaufmann.

\bigskip

\noindent
Badsberg, J. (1992).
\newblock Model search in contingency tables by {CoCo}.
\newblock In Dodge, Y. and Whittaker, J., editors, {\em Computational
  Statistics}, pages 251--256. Physica Verlag, Heidelberg.

\bigskip

\noindent
Becker, S. and LeCun, Y. (1989).
\newblock Improving the convergence of back-propagation learning with second
  order methods.
\newblock In {\em Proceedings of the 1988 Connectionist Models Summer School},
  pages 29--37. Morgan Kaufmann.

\bigskip

\noindent
Beinlich, I., Suermondt, H., Chavez, R., and Cooper, G. (1989).
\newblock The {ALARM} monitoring system: {A} case study with two probabilistic
  inference techniques for belief networks.
\newblock In {\em Proceedings of the Second European Conference on Artificial
  Intelligence in Medicine, {\rm London}}, pages 247--256. Springer Verlag,
  Berlin.

\bigskip

\noindent
Bernardo, J. (1979).
\newblock Expected information as expected utility.
\newblock {\em Annals of Statistics}, 7:686--690.

\bigskip

\noindent
Bernardo, J. and Smith, A. (1994).
\newblock {\em Bayesian Theory}.
\newblock John Wiley and Sons, New York.

\bigskip

\noindent
Buntine, W. (1991).
\newblock Theory refinement on {Bayesian} networks.
\newblock In {\em Proceedings of Seventh Conference on Uncertainty in
  Artificial Intelligence, {\rm Los Angeles, CA}}, pages 52--60. Morgan
  Kaufmann.

\bigskip

\noindent
Buntine, W. (1993).
\newblock Learning classification trees.
\newblock In {\em Artificial Intelligence Frontiers in Statistics: {AI} and
  statistics III}. Chapman and Hall, New York.

\bigskip

\noindent
Buntine, W. (1996).
\newblock A guide to the literature on learning graphical models.
\newblock {\em IEEE Transactions on Knowledge and Data Engineering},
  8:195--210.

\bigskip

\noindent
Chaloner, K. and Duncan, G. (1983).
\newblock Assessment of a beta prior distribution: {PM} elicitation.
\newblock {\em The Statistician}, 32:174--180.

\bigskip

\noindent
Cheeseman, P. and Stutz, J. (1995).
\newblock Bayesian classification {(AutoClass)}: {T}heory and results.
\newblock In Fayyad, U., Piatesky-Shapiro, G., Smyth, P., and Uthurusamy, R.,
  editors, {\em Advances in Knowledge Discovery and Data Mining}, pages
  153--180. AAAI Press, Menlo Park, CA.

\bigskip

\noindent
Chib, S. (1995).
\newblock Marginal likelihood from the {Gibbs} output.
\newblock {\em Journal of the American Statistical Association}, 90:1313--1321.

\bigskip

\noindent
Chickering, D. (1995).
\newblock A transformational characterization of equivalent {Bayesian} network
  structures.
\newblock In {\em Proceedings of Eleventh Conference on Uncertainty in
  Artificial Intelligence, {\rm Montreal, QU}}, pages 87--98. Morgan Kaufmann.

\bigskip

\noindent
Chickering, D. (1996a).
\newblock {\em Learning {Bayesian} Networks from Data}.
\newblock PhD thesis, University of California, Los Angeles, CA.

\bigskip

\noindent
Chickering, D. (1996b).
\newblock Learning equivalence classes of {Bayesian}-network structures.
\newblock In {\em Proceedings of Twelfth Conference on Uncertainty in
  Artificial Intelligence, {\rm Portland, OR}}, pages 87--98. Morgan Kaufmann.

\bigskip

\noindent
Chickering, D., Geiger, D., and Heckerman, D. (1995).
\newblock Learning {Bayesian} networks: Search methods and experimental
  results.
\newblock In {\em Proceedings of Fifth Conference on Artificial Intelligence
  and Statistics, {\rm Ft. Lauderdale, FL}}, pages 112--128. Society for
  Artificial Intelligence in Statistics.

\bigskip

\noindent
Chickering, D. and Heckerman, D. (Revised November, 1996).
\newblock Efficient approximations for the marginal likelihood of incomplete
  data given a {Bayesian} network.
\newblock Technical Report MSR-TR-96-08, Microsoft Research, Redmond, WA.

\bigskip

\noindent
Cooper, G. (1990).
\newblock Computational complexity of probabilistic inference using {Bayesian}
  belief networks ({Research} note).
\newblock {\em Artificial Intelligence}, 42:393--405.

\bigskip

\noindent
Cooper, G. and Herskovits, E. (1992).
\newblock A {Bayesian} method for the induction of probabilistic networks from
  data.
\newblock {\em Machine Learning}, 9:309--347.

\bigskip

\noindent
Cooper, G. and Herskovits, E. (January, 1991).
\newblock A {Bayesian} method for the induction of probabilistic networks from
  data.
\newblock Technical Report SMI-91-1, Section on Medical Informatics, Stanford
  University.

\bigskip

\noindent
Cox, R. (1946).
\newblock Probability, frequency and reasonable expectation.
\newblock {\em American Journal of Physics}, 14:1--13.

\bigskip

\noindent
Dagum, P. and Luby, M. (1993).
\newblock Approximating probabilistic inference in bayesian belief networks is
  np-hard.
\newblock {\em Artificial Intelligence}, 60:141--153.

\bigskip

\noindent
D'Ambrosio, B. (1991).
\newblock Local expression languages for probabilistic dependence.
\newblock In {\em Proceedings of Seventh Conference on Uncertainty in
  Artificial Intelligence, {\rm Los Angeles, CA}}, pages 95--102. Morgan
  Kaufmann.

\bigskip

\noindent
Darwiche, A. and Provan, G. (1996).
\newblock Query {DAGs}: A practical paradigm for implementing belief-network
  inference.
\newblock In {\em Proceedings of Twelfth Conference on Uncertainty in
  Artificial Intelligence, {\rm Portland, OR}}, pages 203--210. Morgan
  Kaufmann.

\bigskip

\noindent
Dawid, P. (1984).
\newblock {S}tatistical theory. {T}he prequential approach (with discussion).
\newblock {\em Journal of the Royal Statistical Society A}, 147:178--292.

\bigskip

\noindent
Dawid, P. (1992).
\newblock Applications of a general propagation algorithm for probabilistic
  expert systmes.
\newblock {\em Statistics and Computing}, 2:25--36.

\bigskip

\noindent
de~Finetti, B. (1970).
\newblock {\em Theory of Probability}.
\newblock Wiley and Sons, New York.

\bigskip

\noindent
Dempster, A., Laird, N., and Rubin, D. (1977).
\newblock Maximum likelihood from incomplete data via the {EM} algorithm.
\newblock {\em Journal of the Royal Statistical Society}, B 39:1--38.

\bigskip

\noindent
DiCiccio, T., Kass, R., Raftery, A., and Wasserman, L. (July, 1995).
\newblock Computing {Bayes} factors by combining simulation and asymptotic
  approximations.
\newblock Technical Report 630, Department of Statistics, Carnegie Mellon
  University, PA.

\bigskip

\noindent
Friedman, J. (1995).
\newblock Introduction to computational learning and statistical prediction.
\newblock Technical report, Department of Statistics, Stanford University.

\bigskip

\noindent
Friedman, J. (1996).
\newblock On bias, variance, 0/1-loss, and the curse of dimensionality.
\newblock {\em Data Mining and Knowledge Discovery}, 1.

\bigskip

\noindent
Friedman, N. and Goldszmidt, M. (1996).
\newblock Building classifiers using {Bayesian} networks.
\newblock In {\em Proceedings {AAAI-96} Thirteenth National Conference on
  Artificial Intelligence, {\rm Portland, OR}}, pages 1277--1284. AAAI Press,
  Menlo Park, CA.

\bigskip

\noindent
Frydenberg, M. (1990).
\newblock The chain graph {Markov} property.
\newblock {\em Scandinavian Journal of Statistics}, 17:333--353.

\bigskip

\noindent
Geiger, D. and Heckerman, D. (Revised February, 1995).
\newblock A characterization of the {Dirichlet} distribution applicable to
  learning {Bayesian} networks.
\newblock Technical Report MSR-TR-94-16, Microsoft Research, Redmond, WA.

\bigskip

\noindent
Geiger, D., Heckerman, D., and Meek, C. (1996).
\newblock Asymptotic model selection for directed networks with hidden
  variables.
\newblock In {\em Proceedings of Twelfth Conference on Uncertainty in
  Artificial Intelligence, {\rm Portland, OR}}, pages 283--290. Morgan
  Kaufmann.

\bigskip

\noindent
Geman, S. and Geman, D. (1984).
\newblock Stochastic relaxation, {Gibbs} distributions and the {Bayesian}
  restoration of images.
\newblock {\em IEEE Transactions on Pattern Analysis and Machine Intelligence},
  6:721--742.

\bigskip

\noindent
Gilks, W., Richardson, S., and Spiegelhalter, D. (1996).
\newblock {\em Markov Chain Monte Carlo in Practice}.
\newblock Chapman and Hall.

\bigskip

\noindent
Good, I. (1950).
\newblock {\em Probability and the Weighing of Evidence}.
\newblock Hafners, New York.

\bigskip

\noindent
Heckerman, D. (1989).
\newblock A tractable algorithm for diagnosing multiple diseases.
\newblock In {\em Proceedings of the Fifth Workshop on Uncertainty in
  Artificial Intelligence, {\rm Windsor, ON}}, pages 174--181. Association for
  Uncertainty in Artificial Intelligence, Mountain View, CA.
\newblock Also in Henrion, M., Shachter, R., Kanal, L., and Lemmer, J.,
  editors, {\em Uncertainty in Artificial Intelligence 5,} pages 163--171.
  North-Holland, New York, 1990.

\bigskip

\noindent
Heckerman, D. (1995).
\newblock A {Bayesian} approach for learning causal networks.
\newblock In {\em Proceedings of Eleventh Conference on Uncertainty in
  Artificial Intelligence, {\rm Montreal, QU}}, pages 285--295. Morgan
  Kaufmann.

\bigskip

\noindent
Heckerman, D. and Geiger, D. (Revised, November, 1996).
\newblock Likelihoods and priors for {Bayesian} networks.
\newblock Technical Report MSR-TR-95-54, Microsoft Research, Redmond, WA.

\bigskip

\noindent
Heckerman, D., Geiger, D., and Chickering, D. (1995a).
\newblock Learning {Bayesian} networks: {T}he combination of knowledge and
  statistical data.
\newblock {\em Machine Learning}, 20:197--243.

\bigskip

\noindent
Heckerman, D., Mamdani, A., and Wellman, M. (1995b).
\newblock Real-world applications of {Bayesian} networks.
\newblock {\em Communications of the ACM}, 38.

\bigskip

\noindent
Heckerman, D. and Shachter, R. (1995).
\newblock Decision-theoretic foundations for causal reasoning.
\newblock {\em Journal of Artificial Intelligence Research}, 3:405--430.

\bigskip

\noindent
H{\o}jsgaard, S., Skj{\o}th, F., and Thiesson, B. (1994).
\newblock User's guide to {BIOFROST}.
\newblock Technical report, Department of Mathematics and Computer Science,
  Aalborg, Denmark.

\bigskip

\noindent
Howard, R. (1970).
\newblock Decision analysis: Perspectives on inference, decision, and
  experimentation.
\newblock {\em Proceedings of the IEEE}, 58:632--643.

\bigskip

\noindent
Howard, R. and Matheson, J. (1981).
\newblock Influence diagrams.
\newblock In Howard, R. and Matheson, J., editors, {\em Readings on the
  Principles and Applications of Decision Analysis}, volume~II, pages 721--762.
  Strategic Decisions Group, Menlo Park, CA.

\bigskip

\noindent
Howard, R. and Matheson, J., editors (1983).
\newblock {\em The Principles and Applications of Decision Analysis}.
\newblock Strategic Decisions Group, Menlo Park, CA.

\bigskip

\noindent
Humphreys, P. and Freedman, D. (1996).
\newblock The grand leap.
\newblock {\em British Journal for the Philosphy of Science}, 47:113--118.

\bigskip

\noindent
Jaakkola, T. and Jordan, M. (1996).
\newblock Computing upper and lower bounds on likelihoods in intractable
  networks.
\newblock In {\em Proceedings of Twelfth Conference on Uncertainty in
  Artificial Intelligence, {\rm Portland, OR}}, pages 340--348. Morgan
  Kaufmann.

\bigskip

\noindent
Jensen, F. (1996).
\newblock {\em An Introduction to {Bayesian} Networks}.
\newblock Springer.

\bigskip

\noindent
Jensen, F. and Andersen, S. (1990).
\newblock Approximations in {Bayesian} belief universes for knowledge based
  systems.
\newblock Technical report, Institute of Electronic Systems, Aalborg
  University, Aalborg, Denmark.

\bigskip

\noindent
Jensen, F., Lauritzen, S., and Olesen, K. (1990).
\newblock Bayesian updating in recursive graphical models by local
  computations.
\newblock {\em Computational Statisticals Quarterly}, 4:269--282.

\bigskip

\noindent
Kass, R. and Raftery, A. (1995).
\newblock Bayes factors.
\newblock {\em Journal of the American Statistical Association}, 90:773--795.

\bigskip

\noindent
Kass, R., Tierney, L., and Kadane, J. (1988).
\newblock Asymptotics in {Bayesian} computation.
\newblock In Bernardo, J., DeGroot, M., Lindley, D., and Smith, A., editors,
  {\em Bayesian Statistics 3}, pages 261--278. Oxford University Press.

\bigskip

\noindent
Koopman, B. (1936).
\newblock On distributions admitting a sufficient statistic.
\newblock {\em Transactions of the American Mathematical Society}, 39:399--409.

\bigskip

\noindent
Korf, R. (1993).
\newblock Linear-space best-first search.
\newblock {\em Artificial Intelligence}, 62:41--78.

\bigskip

\noindent
Lauritzen, S. (1982).
\newblock {\em Lectures on Contingency Tables}.
\newblock University of Aalborg Press, Aalborg, Denmark.

\bigskip

\noindent
Lauritzen, S. (1992).
\newblock Propagation of probabilities, means, and variances in mixed graphical
  association models.
\newblock {\em Journal of the American Statistical Association}, 87:1098--1108.

\bigskip

\noindent
Lauritzen, S. and Spiegelhalter, D. (1988).
\newblock Local computations with probabilities on graphical structures and
  their application to expert systems.
\newblock {\em J. Royal Statistical Society B}, 50:157--224.

\bigskip

\noindent
Lauritzen, S., Thiesson, B., and Spiegelhalter, D. (1994).
\newblock Diagnostic systems created by model selection methods: {A} case
  study.
\newblock In Cheeseman, P. and Oldford, R., editors, {\em AI and Statistics
  IV}, volume Lecture Notes in Statistics, 89, pages 143--152. Springer-Verlag,
  New York.

\bigskip

\noindent
MacKay, D. (1992a).
\newblock Bayesian interpolation.
\newblock {\em Neural Computation}, 4:415--447.

\bigskip

\noindent
MacKay, D. (1992b).
\newblock A practical {Bayesian} framework for backpropagation networks.
\newblock {\em Neural Computation}, 4:448--472.

\bigskip

\noindent
MacKay, D. (1996).
\newblock Choice of basis for the {Laplace} approximation.
\newblock Technical report, Cavendish Laboratory, Cambridge, UK.

\bigskip

\noindent
Madigan, D., Garvin, J., and Raftery, A. (1995).
\newblock Eliciting prior information to enhance the predictive performance of
  {Bayesian} graphical models.
\newblock {\em Communications in Statistics: Theory and Methods},
  24:2271--2292.

\bigskip

\noindent
Madigan, D. and Raftery, A. (1994).
\newblock Model selection and accounting for model uncertainty in graphical
  models using {Occam's} window.
\newblock {\em Journal of the American Statistical Association}, 89:1535--1546.

\bigskip

\noindent
Madigan, D., Raftery, A., Volinsky, C., and Hoeting, J. (1996).
\newblock Bayesian model averaging.
\newblock In {\em Proceedings of the AAAI Workshop on Integrating Multiple
  Learned Models, {\rm Portland, OR}}.

\bigskip

\noindent
Madigan, D. and York, J. (1995).
\newblock Bayesian graphical models for discrete data.
\newblock {\em International Statistical Review}, 63:215--232.

\bigskip

\noindent
Martin, J. and VanLehn, K. (1995).
\newblock Discrete factor analysis: Learning hidden variables in bayesian
  networks.
\newblock Technical report, Department of Computer Science, University of
  Pittsburgh, PA.
\newblock Available at http://bert.cs.pitt.edu/~vanlehn.

\bigskip

\noindent
Meng, X. and Rubin, D. (1991).
\newblock Using {EM} to obtain asymptotic variance-covariance matrices: {The
  SEM} algorithm.
\newblock {\em Journal of the American Statistical Association}, 86:899--909.

\bigskip

\noindent
Neal, R. (1993).
\newblock Probabilistic inference using {Markov} chain {Monte Carlo} methods.
\newblock Technical Report CRG-TR-93-1, Department of Computer Science,
  University of Toronto.

\bigskip

\noindent
Neal, R. (1998).
\newblock Annealed importance sampling.
\newblock arXiv:physics/9803008.

\bigskip

\noindent
Olmsted, S. (1983).
\newblock {\em On representing and solving decision problems}.
\newblock PhD thesis, Department of Engineering-Economic Systems, Stanford
  University.

\bigskip

\noindent
Pearl, J. (1986).
\newblock Fusion, propagation, and structuring in belief networks.
\newblock {\em Artificial Intelligence}, 29:241--288.

\bigskip

\noindent
Pearl, J. (1995).
\newblock Causal diagrams for empirical research.
\newblock {\em Biometrika}, 82:669--710.

\bigskip

\noindent
Pearl, J. and Verma, T. (1991).
\newblock A theory of inferred causation.
\newblock In Allen, J., Fikes, R., and Sandewall, E., editors, {\em Knowledge
  Representation and Reasoning: Proceedings of the Second International
  Conference}, pages 441--452. Morgan Kaufmann, New York.

\bigskip

\noindent
Pitman, E. (1936).
\newblock Sufficient statistics and intrinsic accuracy.
\newblock {\em Proceedings of the Cambridge Philosophy Society}, 32:567--579.

\bigskip

\noindent
Raftery, A. (1995).
\newblock Bayesian model selection in social research.
\newblock In Marsden, P., editor, {\em Sociological Methodology}. Blackwells,
  Cambridge, MA.

\bigskip

\noindent
Raftery, A. (1996).
\newblock {\em Hypothesis testing and model selection}, chapter~10.
\newblock Chapman and Hall.

\bigskip

\noindent
Ramamurthi, K. and Agogino, A. (1988).
\newblock Real time expert system for fault tolerant supervisory control.
\newblock In Tipnis, V. and Patton, E., editors, {\em Computers in
  Engineering}, pages 333--339. American Society of Mechanical Engineers, Corte
  Madera, CA.

\bigskip

\noindent
Ramsey, F. (1931).
\newblock Truth and probability.
\newblock In Braithwaite, R., editor, {\em The Foundations of Mathematics and
  other Logical Essays}. Humanities Press, London.
\newblock Reprinted in Kyburg and Smokler, 1964.

\bigskip

\noindent
Richardson, T. (1997).
\newblock Extensions of undirected and acyclic, directed graphical models.
\newblock In {\em Proceedings of Sixth Conference on Artificial Intelligence
  and Statistics, {\rm Ft. Lauderdale, FL}}, pages 407--419. Society for
  Artificial Intelligence in Statistics.

\bigskip

\noindent
Rissanen, J. (1987).
\newblock Stochastic complexity (with discussion).
\newblock {\em Journal of the Royal Statistical Society, Series B}, 49:223--239
  and 253--265.

\bigskip

\noindent
Robins, J. (1986).
\newblock A new approach to causal interence in mortality studies with
  sustained exposure results.
\newblock {\em Mathematical Modelling}, 7:1393--1512.

\bigskip

\noindent
Rubin, D. (1978).
\newblock Bayesian inference for causal effects: {The} role of randomization.
\newblock {\em Annals of Statistics}, 6:34--58.

\bigskip

\noindent
Russell, S., Binder, J., Koller, D., and Kanazawa, K. (1995).
\newblock Local learning in probabilistic networks with hidden variables.
\newblock In {\em Proceedings of the Fourteenth International Joint Conference
  on Artificial Intelligence, {\rm Montreal, QU}}, pages 1146--1152. Morgan
  Kaufmann, San Mateo, CA.

\bigskip

\noindent
Saul, L., Jaakkola, T., and Jordan, M. (1996).
\newblock Mean field theory for sigmoid belief networks.
\newblock {\em Journal of Artificial Intelligence Research}, 4:61--76.

\bigskip

\noindent
Savage, L. (1954).
\newblock {\em The Foundations of Statistics}.
\newblock Dover, New York.

\bigskip

\noindent
Schervish, M. (1995).
\newblock {\em Theory of Statistics}.
\newblock Springer-Verlag.

\bigskip

\noindent
Schwarz, G. (1978).
\newblock Estimating the dimension of a model.
\newblock {\em Annals of Statistics}, 6:461--464.

\bigskip

\noindent
Sewell, W. and Shah, V. (1968).
\newblock Social class, parental encouragement, and educational aspirations.
\newblock {\em American Journal of Sociology}, 73:559--572.

\bigskip

\noindent
Shachter, R. (1988).
\newblock Probabilistic inference and influence diagrams.
\newblock {\em Operations Research}, 36:589--604.

\bigskip

\noindent
Shachter, R., Andersen, S., and Poh, K. (1990).
\newblock Directed reduction algorithms and decomposable graphs.
\newblock In {\em Proceedings of the Sixth Conference on Uncertainty in
  Artificial Intelligence, {\rm Boston, MA}}, pages 237--244. Association for
  Uncertainty in Artificial Intelligence, Mountain View, CA.

\bigskip

\noindent
Shachter, R. and Kenley, C. (1989).
\newblock Gaussian influence diagrams.
\newblock {\em Management Science}, 35:527--550.

\bigskip

\noindent
Silverman, B. (1986).
\newblock {\em Density Estimation for Statistics and Data Analysis}.
\newblock Chapman and Hall, New York.

\bigskip

\noindent
Singh, M. and Provan, G. (November, 1995).
\newblock Efficient learning of selective {Bayesian} network classifiers.
\newblock Technical Report MS-CIS-95-36, Computer and Information Science
  Department, University of Pennsylvania, Philadelphia, PA.

\bigskip

\noindent
Spetzler, C. and Stael~von Holstein, C. (1975).
\newblock Probability encoding in decision analysis.
\newblock {\em Management Science}, 22:340--358.

\bigskip

\noindent
Spiegelhalter, D., Dawid, A., Lauritzen, S., and Cowell, R. (1993).
\newblock Bayesian analysis in expert systems.
\newblock {\em Statistical Science}, 8:219--282.

\bigskip

\noindent
Spiegelhalter, D. and Lauritzen, S. (1990).
\newblock Sequential updating of conditional probabilities on directed
  graphical structures.
\newblock {\em Networks}, 20:579--605.

\bigskip

\noindent
Spirtes, P., Glymour, C., and Scheines, R. (1993).
\newblock {\em Causation, Prediction, and Search}.
\newblock Springer-Verlag, New York.

\bigskip

\noindent
Spirtes, P. and Meek, C. (1995).
\newblock Learning {Bayesian} networks with discrete variables from data.
\newblock In {\em Proceedings of First International Conference on Knowledge
  Discovery and Data Mining, {\rm Montreal, QU}}, pages 294--299. Morgan
  Kaufmann.

\bigskip

\noindent
Suermondt, H. and Cooper, G. (1991).
\newblock A combination of exact algorithms for inference on {Bayesian} belief
  networks.
\newblock {\em International Journal of Approximate Reasoning}, 5:521--542.

\bigskip

\noindent
Thiesson, B. (1995a).
\newblock Accelerated quantification of {Bayesian} networks with incomplete
  data.
\newblock In {\em Proceedings of First International Conference on Knowledge
  Discovery and Data Mining, {\rm Montreal, QU}}, pages 306--311. Morgan
  Kaufmann.

\bigskip

\noindent
Thiesson, B. (1995b).
\newblock Score and information for recursive exponential models with
  incomplete data.
\newblock Technical report, Institute of Electronic Systems, Aalborg
  University, Aalborg, Denmark.

\bigskip

\noindent
Thomas, A., Spiegelhalter, D., and Gilks, W. (1992).
\newblock Bugs: A program to perform {Bayesian} inference using {Gibbs}
  sampling.
\newblock In Bernardo, J., Berger, J., Dawid, A., and Smith, A., editors, {\em
  Bayesian Statistics 4}, pages 837--842. Oxford University Press.

\bigskip

\noindent
Tukey, J. (1977).
\newblock {\em Exploratory Data Analysis}.
\newblock Addison--Wesley.

\bigskip

\noindent
Tversky, A. and Kahneman, D. (1974).
\newblock Judgment under uncertainty: Heuristics and biases.
\newblock {\em Science}, 185:1124--1131.

\bigskip

\noindent
Verma, T. and Pearl, J. (1990).
\newblock Equivalence and synthesis of causal models.
\newblock In {\em Proceedings of Sixth Conference on Uncertainty in Artificial
  Intelligence, {\rm Boston, MA}}, pages 220--227. Morgan Kaufmann.

\bigskip

\noindent
Whittaker, J. (1990).
\newblock {\em Graphical Models in Applied Multivariate Statistics}.
\newblock John Wiley and Sons.

\bigskip

\noindent
Winkler, R. (1967).
\newblock The assessment of prior distributions in {Bayesian} analysis.
\newblock {\em American Statistical Association Journal}, 62:776--800.

\trsj{}{}{\end{article}}{}

\end{document}